\documentclass[11pt]{article}
\usepackage[margin=1in, centering]{geometry}
\usepackage{graphicx} 
\usepackage{amsmath,amssymb,amsthm,mathtools}
\usepackage{natbib}
\usepackage[colorlinks=true, citecolor=blue]{hyperref}
\usepackage{xcolor}

\newtheorem{theorem}{Theorem}[section]
\newtheorem{corollary}[theorem]{Corollary}
\newtheorem{lemma}[theorem]{Lemma}
\newtheorem{fact}[theorem]{Fact}

\newtheorem{definition}[theorem]{Definition}
\theoremstyle{remark}

\usepackage[linesnumbered,ruled]{algorithm2e}
\SetKwInput{KwInput}{Input}
\SetKwInput{KwOutput}{Output}
\newcommand{\A}{\mathcal{A}}
\newcommand{\C}{\mathcal{C}}
\newcommand{\CC}{\mathcal{C}^\cap_\infty}
\newcommand{\F}{\mathcal{F}}
\newcommand{\G}{\mathcal{G}}
\newcommand{\M}{\mathcal{M}}
\newcommand{\T}{\mathcal{T}}
\newcommand{\V}{\mathcal{V}}
\newcommand{\X}{\mathcal{X}}
\newcommand{\h}{h_{\mathrm{fin}}}
\newcommand{\cd}{\mathsf{CDim}}
\newcommand{\ld}{\mathsf{Ldim}}
\newcommand{\nbd}{\mathsf{NBDim}}
\newcommand{\core}{\mathsf{core}}
\newcommand{\profile}{\mathsf{profile}}

\title{Private Uniform Language Generation}
\author{Di Wang\thanks{King Abdullah University of Science and Technology. \texttt{di.wang@kaust.edu.sa}} \and Peng Ye\thanks{King Abdullah University of Science and Technology. \texttt{peng.ye@kaust.edu.sa}}}
\date{}

\begin{document}

\maketitle

\begin{abstract}
    We study uniform language generation under approximate differential privacy. Given a uniformly bounded number of distinct examples from an unknown target language, a generator must produce infinitely many new examples from that language with probability one while protecting the privacy of its input. We introduce a new combinatorial dimension termed the nested branching dimension and prove that a language class is privately uniformly generatable if and only if both its closure dimension and nested branching dimension are finite. Previous work characterizes non-private uniform generation by the finiteness of the closure dimension alone. Our results therefore identify an additional combinatorial obstruction imposed by privacy.
\end{abstract}
\newpage

\section{Introduction}

Motivated in part by the remarkable success of large language models, recent work has studied the problem of language generation~\cite{kleinberg2024language}, in which an algorithm observes examples from an unknown language and seeks to produce new examples from that language. This objective differs fundamentally from classical predictive learning, where the goal is to make accurate predictions on unseen instances after observing examples labeled by an unknown target.

A central goal in learning theory is to characterize which classes are learnable in terms of combinatorial parameters. Canonical examples include the VC dimension~\cite{vapnik1971uniform}, which characterizes PAC learnability, and the Littlestone dimension~\cite{littlestone1988learning}, which characterizes online learnability. Such characterizations do more than establish the existence of learning algorithms: they identify the structural properties that govern learnability and often yield quantitative bounds on the amount of data required.

Such combinatorial characterization has also been identified for a language generation task called uniform language generation. Let $\C$ be a collection of languages, each of which is an infinite subset of a countable universe. We say $\C$ is uniformly generatable if there exists an algorithm and a finite $n$ such that for every language $L\in\C$, it takes as input a dataset $S$ of $n$ distinct examples from $L$ and produces an infinite subset of $L\setminus S$. Here, ``uniform'' means $n$ only depends on the entire class $\C$ and is the same for every $L\in\C$. Raman, Li, and Tewari~\cite{raman2025generation} give a complete combinatorial characterization of this property in terms of a parameter called the closure dimension, and further show that uniform generatability is incomparable with both PAC and online learnability. Thus, although learning and generation admit parallel characterization questions, the structural conditions governing the two tasks are fundamentally different.

The examples observed by a generator, however, may contain sensitive information, raising the question of whether generation can be carried out while protecting the privacy of the input data. Differential privacy (DP)~\cite{dwork2006calibrating} provides a strong and widely used formalization of such protection: informally, it requires that changing a single example in the input dataset has only a limited effect on the distribution of the algorithm's output.

Imposing differential privacy can fundamentally change the conditions that determine whether a task is possible. This phenomenon is well known in learning theory: ordinary PAC learnability is characterized by the VC dimension, whereas (approximate) private PAC learnability is characterized by the Littlestone dimension~\cite{alon2019private,bun2020equivalence}. These two dimensions can differ dramatically for the same class: for example, the class of thresholds over $\mathbb{N}$ has VC dimension $1$ but infinite Littlestone dimension, and is therefore PAC learnable but not privately PAC learnable. Thus, the classes that are learnable in the private and non-private settings can differ.

This naturally motivates the corresponding question for generation. Suppose a generator observes a uniformly bounded number of distinct examples from an unknown target language $L\in \C$. Based on these examples, it must generate infinitely many new examples in $L$ with probability $1$, while satisfying differential privacy with respect to its input sample. In the non-private setting, the closure dimension completely characterizes when uniform generation is possible. Does this characterization continue to hold under differential privacy, perhaps with a larger sample size, or does privacy fundamentally change the structure governing generation? This leads to our main question:

\begin{center}
\emph{Which classes are uniformly generatable under differential privacy?}
\end{center}

\subsection{Main Results}
We assume a countable example space, e.g., the set of all strings of finite length over a finite alphabet. A language is an infinite subset of the example space, and a language class is a (possibly uncountable) collection of languages. In this work, we focus on approximate differential privacy because any algorithm satisfying pure differential privacy must admit a nonzero failure probability and cannot succeed almost surely. To characterize which language classes are uniformly generatable under differential privacy, we introduce a new combinatorial dimension, which we call the nested branching dimension (Section~\ref{sec:dimension}). Our main result gives a complete characterization in terms of the closure dimension and the nested branching dimension.

\begin{theorem}
    A language class $\C$ is uniformly generatable under approximate differential privacy if and only if both its closure dimension and nested branching dimension are finite.
\end{theorem}

This characterization differs from the non-private setting, where the finiteness of the closure dimension alone characterizes uniform generatability~\cite{raman2025generation}. Moreover, the two finiteness conditions above are incomparable: there exist classes with a finite closure dimension and an infinite nested branching dimension, as well as classes with an infinite closure dimension and a finite nested branching dimension. Thus, both conditions in the characterization are necessary, demonstrating that privacy introduces an additional combinatorial obstruction to uniform generation.

Specifically, our characterization follows from the following quantitative upper and lower bounds on the sample complexity of private uniform generation.

\begin{theorem}
    Let $\C$ be a language class with finite closure dimension $\cd(\C)$ and nested branching dimension $\nbd(\C)$. For any $\varepsilon, \delta\in(0,1)$, there exists an $(\varepsilon,\delta)$-differentially private uniform generator using
    \begin{equation*}
        \cd(\C) + 1 + O\left(\frac{\nbd(\C)\cdot 2^{\nbd(\C)}}{\varepsilon}\left(\nbd(\C) + \log \left(\frac{1}{\varepsilon\delta}\right)\right)\right)
    \end{equation*}
    distinct examples.
\end{theorem}

\begin{theorem}
    Let $n\ge1$, $0<\varepsilon\le1$, and $0\le\delta\le1/(12n)$. If there is an $(\varepsilon,\delta)$-differentially private uniform generator for language class $\C$ using $n$ distinct examples, then
    \begin{equation*}
        n\ge\max\left(\cd(\C)+1,\Omega\left(\log^*\nbd(\C)\right)\right).
    \end{equation*}
\end{theorem}
The $\cd(\C) + 1$ term in the two bounds is the sample complexity of non-private uniform generation. Thus, the additional term reflects the cost of privacy.

\subsection{Related Work}

\paragraph*{Language Generation.}
Building on the Gold--Angluin framework of language identification in the limit~\cite{gold1967language,angluin1980inductive}, Kleinberg and Mullainathan~\cite{kleinberg2024language} introduced language generation in the limit. This problem and its variants have since been studied in a growing body of work (e.g., generation with breadth and density guarantees~\cite{charikar2025exploring,kalavasis2025limits,kalavasis2026characterizations,kleinberg2025density}, generation from noisy examples~\cite{raman2025noisy,bai2026language,mehrotra2026infinite,li2026noise}, and generation with feedback~\cite{charikar2025exploring,bai2026language,hanneke2026feedback}). Uniform generation, the focus of our work, was also proposed by Kleinberg and Mullainathan, who established algorithms for finite collections. Raman, Li, and Tewari~\cite{raman2025generation} subsequently characterized uniform generatability by the finiteness of the closure dimension. More recently, private language generation has been studied under continual release~\cite{mehrotra2026differentially} and in the agnostic statistical setting~\cite{li2026price}. For finite collections, Mehrotra, Velegkas, Yu, and Zhou~\cite{mehrotra2026differentially} establish uniform bounds under pure DP with a nonzero failure probability. Our work characterizes almost sure uniform generation under approximate DP for potentially uncountable collections.

\paragraph*{Private Learning.}
Differentially private PAC learning was introduced in~\cite{kasiviswanathan2011can}. Subsequent work established combinatorial characterizations of private learnability. Under pure DP, the probabilistic representation dimension characterizes learnability and governs sample complexity~\cite{beimel2019characterizing}. Under approximate DP, private PAC learnability is characterized by finiteness of the Littlestone dimension~\cite{alon2019private,bun2020equivalence}. Our characterization introduces a different parameter, the nested branching dimension. Nevertheless, our lower bound connects private generation to private learning through the lower bound for the interior point problem, whose sample complexity is equivalent, up to constant factors for fixed accuracy parameters, to that of private proper PAC learning of thresholds~\cite{bun2015differentially}.

\paragraph*{Characterization of Learnability.}
Classical characterizations include the VC dimension for binary PAC learning~\cite{blumer1989learnability} and the Littlestone dimension for online learning~\cite{littlestone1988learning}. Further examples include the DS dimension for multiclass PAC learning~\cite{brukhim2022characterization} and the fat-shattering dimension for learning real-valued functions~\cite{bartlett1996fat}. However, such characterizations do not always exist, as demonstrated by EMX learning~\cite{bendavid2019learnability} and distribution learning~\cite{lechner2024inherent}.

\section{Technical Overview}
\label{sec:technical_overview}

We begin by reviewing the non-private characterization of uniform generation. For a set $S$ of distinct examples from an unknown target language in $\C$, define its closure by 
\begin{equation*}
    \langle S\rangle_\C=\bigcap_{L\in\C:\,S\subseteq L}L.
\end{equation*}
This is the set of points contained in every target language consistent with $S$. Since the example space $\X$ is countable, any successful uniform generator must almost surely output an infinite subset of this closure. Successful generation therefore requires sufficiently many distinct examples to ensure that the closure is infinite. The closure dimension $\cd(\C)$ measures the largest size of a realizable sample whose closure is finite. By definition, $\langle S\rangle_\C$ is infinite whenever $\lvert S\rvert>\cd(\C)$. Once this condition holds, a non-private generator can output $\langle S\rangle_\C\setminus S$, obtaining infinitely many new examples from the target language. However, the closure may change substantially when a single example is replaced, motivating an additional structural condition that determines when we can extract a stable infinite subset.

\paragraph*{Nested branching dimension.}
We introduce nested branching dimension to capture a structural obstruction to private uniform generation. A nested branching tree is a rooted tree (with possibly infinitely many children at each node) whose nodes are labeled by sets in $\CC$, the family of infinite sets obtained as intersections of languages from $\C$. In particular, this family contains every infinite closure $\langle S\rangle_\C$. Each child label is contained in its parent label, and the intersection of all child labels at every internal node is finite. Thus, each child label permits an infinite output, but no infinite output is valid for all children simultaneously. The nested branching dimension $\nbd(\C)$ is the maximum depth of such a tree. We also define the rank $\rho(K)$ of $K\in\CC$ as the maximum depth of such a tree rooted at $K$.

\subsection{Upper Bound}

Our algorithm first constructs a nested sequence of closures and then extracts a subset (which we call an iterated core) from each closure. We prove that the resulting sequence forms a bounded number of contiguous blocks whose stability across neighboring datasets enables private selection. We begin by developing the rank and core properties.

\paragraph*{Rank cores.}
For $K\in\CC$, define its rank core by
\begin{equation*}
    \core(K)\coloneqq\bigcap_{T\in\CC:\,T\subseteq K,\ \rho(T)=\rho(K)}T.
\end{equation*}
Nested branching guarantees that this core is infinite. Otherwise, trees of depth $\rho(K)$ rooted at the sets in the intersection could be attached beneath $K$ to obtain a deeper tree. For nested sets $H\subseteq K$ of equal rank, if taking the core of $K$ preserves its rank, then $\core(H)=\core(K)$. Thus, along a nested sequence of equal-rank sets, a core that preserves rank is shared by all subsequent sets. When this agreement persists across neighboring samples, a stability-based mechanism can privately output the common core. However, taking the core may decrease rank. Therefore, equality of rank alone does not guarantee that nested sets have the same core.

\paragraph*{Profiles and iterated cores.}
We refine rank by repeatedly applying the core operation. Each iteration either strictly decreases rank or produces a fixed point, so the process terminates. The profile $\profile(K)$ records the trajectory of ranks. The iterated core $\core^*(K)$ is the final fixed set, which remains an infinite subset of $K$. Each profile is a strictly decreasing sequence of positive integers bounded by $\nbd(\C)$. There are therefore at most $2^{\nbd(\C)}-1$ possible profiles. We equip profiles with a total order $\preceq$ and establish the following properties for $H,K\in\CC$ with $H\subseteq K$:
\begin{equation*}
    \begin{aligned}
        \profile(K)&\preceq\profile(H),\\
        \profile(H)=\profile(K)
        &\quad\Longrightarrow\quad
        \core^*(H)=\core^*(K).
    \end{aligned}
\end{equation*}
Consequently, any nested sequence can be partitioned into at most $2^{\nbd(\C)}-1$ contiguous blocks, each with a common iterated core. Our algorithm will construct such a sequence from the sample and privately output the common iterated core of a selected block.

\paragraph*{Robust closures.}
We construct the desired nested sequence by intersecting closures obtained after discarding examples from the sample. For a dataset $S$, define the robust closure by
\begin{equation*}
    Q_t(S)\coloneqq\bigcap_{\substack{
        T\subseteq\hat S:\,\lvert T\rvert=\lvert \hat S\rvert-t\\
        T\subseteq L\text{ for some }L\in\C
    }}\langle T\rangle_\C,
\end{equation*}
where $\hat S$ is the set of distinct examples in $S$ (to ensure the sequence is well-defined and nested on all inputs, we set $Q_t(S)=\X$ when $t$ is small and no $T$ satisfies the condition, and set $Q_t(S)=\varnothing$ when $t$ is too large, see Definition~\ref{def:robust_closure} for the formal definition). The resulting sequence is nested: $Q_0(S)\supseteq Q_1(S)\supseteq\cdots$. Moreover, the sequences on any two neighboring datasets $S,S'$ satisfy the following interlacing relations:
\begin{equation*}
    Q_{t+1}(S)\subseteq Q_t(S'),
    \qquad
    Q_{t+1}(S')\subseteq Q_t(S).
\end{equation*}
This interlacing allows us to control how the profile blocks change under a replacement. However, $Q_t(S)$ may be finite and cannot provide an infinite output. We need to ensure that a sample from the target language with sufficiently many distinct examples yields a long sequence of infinite robust closures.

\paragraph*{Constructing a long sequence of infinite robust closures.}
To obtain a long sequence of infinite robust closures, we introduce the finite-intersection Helly number. Define the finite-intersection Helly number $\h(\C)$  as the smallest positive integer $h$ such that every nonempty family of languages from $\C$ with a finite intersection has a subfamily of at most $h$ languages whose intersection is also finite. We show that
\begin{equation*}
    \h(\CC)=\h(\C)\le\nbd(\C)+1.
\end{equation*}
For a sample $S$ from the target language, $Q_t(S)$ is infinite whenever $\lvert\hat S\rvert-\h(\C)t>\cd(\C)$. Indeed, any collection of at most $\h(\C)$ subsets obtained by discarding $t$ examples shares more than $\cd(\C)$ examples. The closure of these common examples is infinite and lies in all the corresponding closures. Therefore, every such collection of closures has an infinite intersection. The Helly property then ensures that the intersection of all closures defining $Q_t(S)$ is infinite. We can therefore obtain a nested sequence of
\begin{equation*}
    \left\lceil\frac{\lvert\hat S\rvert-\cd(\C)}{\h(\C)}\right\rceil
    \ge
    \left\lceil\frac{\lvert\hat S\rvert-\cd(\C)}{\nbd(\C)+1}\right\rceil
\end{equation*}
infinite robust closures, all contained in the target language.

\paragraph*{Private selection.}
Every robust closure $Q_t(S)$ in the sequence constructed above is infinite and therefore a member of $\CC$. Thus, $\profile(Q_t(S))$ is well-defined. Since the sequence is nested, those $Q_t(S)$ with the same profile have the same iterated core. Consequently, each profile $\sigma$ occupies a contiguous block in the sequence with length $\lambda_S(\sigma)$ and a common iterated core $\gamma_S(\sigma)$. Interlacing ensures that each block endpoint moves by at most one under a replacement. Thus, for neighboring datasets $S,S'$, we have $\lvert\lambda_S(\sigma)-\lambda_{S'}(\sigma)\rvert\le2$ and
\begin{equation*}
    \lambda_S(\sigma)>2 \quad\Longrightarrow\quad
    \lambda_{S'}(\sigma)>0 \ \text{and}\ \gamma_S(\sigma)=\gamma_{S'}(\sigma).
\end{equation*}
The algorithm selects an occurring profile $\sigma$ with probability proportional to $\exp(\varepsilon\lambda_S(\sigma)/8)$ and returns $\gamma_S(\sigma)$. 

When a profile $\sigma$ occurs in both neighboring datasets with the same associated output, its exponential weights have a ratio of at most $e^{\varepsilon/4}$. Otherwise, its block length is at most two. Since there are $O(2^{\nbd(\C)})$ profile blocks, a sufficiently long sequence ensures that at least one block is long. Its exponential weight makes the combined contribution of profiles that disappear or have a different associated output in the neighboring dataset a sufficiently small fraction of the total weight. This bounds their total probability by $\delta$ and allows us to control the ratio of the normalizing factors, yielding $(\varepsilon,\delta)$-differential privacy. Every possible choice on a sample from the target language with sufficiently many distinct examples yields an infinite subset of the target, so this utility holds with probability one.

\paragraph*{Excluding input examples.}
The selected infinite set may still contain input examples. To exclude them from the selected output, we partition the set into $b$ infinite parts, choose one uniformly at random, and remove the input examples from that part. For a fixed selected set, neighboring datasets produce the same output unless the chosen part contains one of the at most two examples changed by a replacement. This event has a probability of at most $2/b$. Taking $b$ sufficiently large makes the additional contribution to $\delta$ arbitrarily small, while the output remains infinite and excludes all input examples.

\subsection{Lower Bound}

To prove the lower bound, suppose an $(\varepsilon,\delta)$-differentially private uniform generator succeeds after observing $n$ distinct examples. The non-private characterization already gives $n>\cd(\C)$. To obtain the additional bound imposed by privacy, we first derive a finite-intersection property and then use it to reduce nested branching to the private interior-point problem.

\paragraph*{Generatability implies Helly property.}
We first show that $\h(\C)\le n$ whenever $\delta<1/n$. If this bound fails, there is a family of languages whose intersection is finite, while every subfamily of at most $n$ languages has an infinite intersection. We consider two cases.

First, suppose some finite subfamily has a finite intersection and choose one of minimum size. Intersecting all but one of its languages and removing their common finite intersection yields pairwise disjoint infinite regions. Encoding records by distinct points in these regions turns the generator into a private mechanism that must select a record value present in its input. A classical packing argument then gives $1\le n\delta$, contradicting $\delta<1/n$.

Second, suppose every finite subfamily has an infinite intersection. For arbitrary $N$ and $\eta>0$, we construct a nested sequence $B_0\supseteq\cdots\supseteq B_N$, each being a finite intersection of languages from the family. A record with value $i$ at position $t$ is encoded by a point $a_{i,t}\in B_{i-1}$, with all encoding points distinct. At step $i$, we intersect $B_{i-1}$ with finitely many additional languages to obtain $B_i$, ensuring that the output fails containment in $B_i$ with a probability of at least $1-\eta$ on every input encoded using values at most $i$. This can be achieved because the full intersection is finite, $\X$ is countable, and there are only finitely many such encoded inputs. Each $B_i$ remains infinite by assumption. Nesting ensures that the first failed containment cannot precede the minimum input value. By construction, it occurs at or before the maximum input value with a probability of at least $1-\eta$. This yields a private interior-point mechanism over arbitrarily large finite domains with an arbitrarily small failure probability, contradicting the lower bound for this problem~\cite{bun2015differentially}.

\paragraph*{Reduction to binary branching.}
We show that
\begin{equation*}
    \nbd(\C)\le 2^{\nbd_2(\C)}-1,
\end{equation*}
where $\nbd_2(\C)$ denotes the maximum depth of a binary nested branching tree.

We prove this by induction on binary rank. If a node has infinitely many children, the Helly bound ensures that we can retain at most $n$ children whose labels still have finite intersection. This pruning preserves depth and provides a finite collection from which we can choose a minimal subcollection with finite intersection. Minimality allows us to form binary splits by separating one label from the intersection of the remaining labels. If the tree is too deep for its root's binary rank, we iterate over the finitely many children in this minimal subcollection and apply the induction hypothesis to find two subsets of the root label with a finite intersection, each having the same binary rank as the root. Joining binary trees rooted at these sets beneath the root adds one binary level, contradicting the root's binary rank.

\paragraph*{From binary branching to interior point.}
We prove that a private uniform generator yields a private interior-point mechanism on an ordered domain of size $\Theta(\nbd_2(\C))$. The interior-point lower bound and the rank comparison give
\begin{equation*}
    n=\Omega\!\left(\log^*\nbd_2(\C)\right)
    =\Omega\!\left(\log^*\nbd(\C)\right).
\end{equation*}

Following the idea of~\cite{bun2015differentially}, we first choose a reference leaf uniformly at random. Encode each of the $n-1$ records $x_i$ by a leaf that follows the reference path for $x_i$ levels and then continues independently at random. We choose fixed distinct representative points for each leaf and input position, avoiding finite intersections between sibling labels. The representatives from the encoded leaves and the reference leaf form the generator's input. We decode the generated set by returning the deepest level on the reference path whose label contains it. All input points lie in the label at level $\min_i x_i$, so the generated set is contained in this label almost surely. Thus, the decoded value is at least the minimum.

For the maximum, beyond $\max_i x_i$, the encoded continuations are independent of the random reference path. Privacy limits the generator's ability to predict the next branch beyond $\max_i x_i$ along the random reference path. \cite{bun2015differentially} makes the probability of a correct guess small by using a large digit base. However, each node has only two children here, giving a base of size only $2$. We address this by grouping a constant number of consecutive binary edges into one step, obtaining sufficiently many branch choices while reducing depth by only a constant factor. This makes the decoded value unlikely to exceed the maximum and gives the required interior point guarantee.

\section{Preliminaries}

Let $\X$ denote a countable example space. A language $L$ is an infinite subset of $\X$. A language class $\C\subseteq\mathcal P(\X)$ is a nonempty, possibly uncountable family of languages. For a nonempty family $\F\subseteq\mathcal P(\X)$, we use $\bigcap\F$ to denote the intersection of all sets in $\F$. For a sequence $S=(x_1,\dots,x_n)\in\X^n$, write $\hat S=\{x_1,\dots,x_n\}$ for the set of its distinct entries.
For a positive integer $k$, write $[k]=\{1,\ldots,k\}$. We use $\log$ for the base-$2$ logarithm. For $x\ge1$, the iterated logarithm $\log^*x$ is the number of applications of $\log$ needed to obtain a value of at most $1$.

\subsection{Uniform Language Generation}

\begin{definition}[Uniform Language Generation~\cite{kleinberg2024language}]
\label{def:uniform_language_generation}
    A language class $\C$ is uniformly generatable if there exists a randomized algorithm $\A$ with finite sample size $n$ such that for every $L\in\C$ and every sequence $S=(x_1,\dots,x_n)\in L^n$ with $n$ distinct elements, we have
    \begin{equation*}
        \Pr[\lvert\A(S)\rvert = \infty\text{ and }\A(S)\subseteq L\setminus\{x_1,\dots,x_n\}] = 1.
    \end{equation*}
    Such an algorithm is called a uniform generator for $\C$.
\end{definition}

For a non-empty finite set $T$ contained in some language from $\C$, define the closure of $T$ as
\begin{equation*}
    \langle T\rangle_\C
    \coloneqq\bigcap_{L\in\C,\,T\subseteq L}L.
\end{equation*}
The closure has the following monotonicity property: if $T\subseteq T'$ and $T'$ is contained in some language from $\C$, then $\langle T\rangle_\C\subseteq\langle T'\rangle_\C$.

The closure dimension quantifies when closures must be infinite.

\begin{definition}[Closure Dimension~\cite{raman2025generation}]
    The closure dimension of a language class $\C$, denoted by $\cd(\C)$, is defined as the largest integer $d$ for which there exists a nonempty finite set $T$ contained in some $L\in\C$ such that $\lvert\langle T\rangle_\C\rvert <\infty$ and $\lvert T\rvert \ge d$. If this holds for arbitrarily large $d$, then $\cd(\C) = \infty$. If this is not true even for $d = 1$ and $\lvert \bigcap \C\rvert<\infty$, then $\cd(\C) = 0$. Otherwise $\cd(\C) = -1$.
\end{definition}

Raman, Li, and Tewari~\cite{raman2025generation} show that uniform generatability is characterized by finiteness of the closure dimension. While their work focuses on deterministic algorithms, their argument naturally extends to our randomized definition through the countability of $\X$ (see Lemma~\ref{lem:lower_closure_dimension} for a formal proof). We also remark that they actually consider a slightly weaker model, in which the algorithm still receives an infinite sequence of valid examples after observing $S$ and outputs a new example at each round rather than producing an infinite subset all at once. However, their algorithm works in the stronger model we study.

\begin{theorem}[\cite{raman2025generation}]
\label{thm:uniform_closure_dimension}
    A language class $\C$ is uniformly generatable if and only if it has finite closure dimension.
\end{theorem}

\subsection{Differential Privacy}

\begin{definition}[Differential Privacy~\cite{dwork2006calibrating}]
    A randomized algorithm $\A$ is $(\varepsilon,\delta)$-differentially private if for any two datasets $S$ and $S'$ that differ in one element and any measurable subset $O$ of its output range, we have
    \begin{equation*}
        \Pr[\A(S)\in O]\le e^\varepsilon\Pr[\A(S')\in O] +\delta.
    \end{equation*}
\end{definition}

The privacy guarantee applies to all datasets in $\X^n$, including those with repeated entries and those whose support is not contained in any language in $\C$. The uniform generation guarantee is required only for datasets consisting of $n$ distinct examples from a target language.

Following the convention of using a constant privacy parameter in private learning~\cite{alon2022private}, we fix $\varepsilon=1$ and require $\delta$ to vanish faster than $1/n$, where $n$ is the input sample size. A language class $\C$ is privately uniformly generatable if there exist an integer $n_0\ge1$, a family of randomized algorithms $(\A_n)_{n\ge n_0}$, and parameters $\delta_n=o(1/n)$ such that, for every $n\ge n_0$, $\A_n$ is $(1,\delta_n)$-differentially private on all of $\X^n$ and satisfies the uniform generation guarantee in Definition~\ref{def:uniform_language_generation} for every sequence of $n$ distinct examples from every $L\in\C$.

\section{Nested Branching and the Helly Property}
\label{sec:dimension}

In this section, we formally define the nested branching dimension and the finite-intersection Helly property. Let $\C$ be a language class. Define its infinite-intersection closure by
\begin{equation*}
    \CC \coloneqq\left\{\bigcap\F:\varnothing\neq\F\subseteq\C,\ \left\lvert\bigcap\F\right\rvert=\infty\right\}.
\end{equation*}

\begin{definition}[Nested Branching Tree]
    A nested branching tree for $\C$ of depth $d$ is a rooted tree with the following properties:
    \begin{enumerate}
        \item Every node $u$ is labeled by some $K_u\in \CC$.
        \item Every root-to-leaf path has a length of $d$, i.e., it contains exactly $d$ nodes.
        \item If $u$ is the parent of $v$, then $K_v\subseteq K_u$.
        \item For every internal node $u$, we have $\lvert \bigcap_{v\text{ is a child of }u} K_v\rvert < \infty$.
    \end{enumerate}
    Moreover, if each of its internal nodes has at most $k$ children, then it is called a $k$-ary nested branching tree.
\end{definition}

\begin{definition}[Rank]
    For every $K\in\CC$, let $D(\C, K)$ be the set of all $d$ such that there exists a nested branching tree of depth $d$ whose root is labeled by $K$. The rank of $K$ is defined as
    \begin{equation*}
        \rho(\C, K) \coloneqq \sup D(\C, K).
    \end{equation*}
    
    Similarly, let $D_k(\C, K)$ be the set of all $d$ such that there exists a $k$-ary nested branching tree for $\C$ of depth $d$ whose root is labeled by $K$. The $k$-rank of $K$ is defined as
    \begin{equation*}
        \rho_k(\C, K) \coloneqq \sup D_k(\C, K).
    \end{equation*}

    When the class $\C$ is clear from context, we may omit it and write $\rho(K)$ and $\rho_k(K)$.
\end{definition}

The rank has the following monotonicity property.

\begin{fact}
\label{fact:upper_rank_monotonicity}
    If $H,K\in\CC$ and $H\subseteq K$, then $\rho(H)\le\rho(K)$ and $\rho_k(H)\le \rho_k(K)$ for every $k\ge 2$.
\end{fact}
\begin{proof}
    Given a nested branching tree rooted at $H$, replace its root label by $K$. Every child of the root is still labeled by a subset of the root label, and all other conditions are unchanged. The resulting tree has the same depth and is rooted at $K$.
\end{proof}

The nested branching dimension measures how deep a nested branching tree can be.

\begin{definition}[Nested Branching Dimension]
    The nested branching dimension of $\C$ is defined as
    \begin{equation*}
        \nbd(\C) \coloneqq \sup \{\rho(\C, K):K\in\CC\}.
    \end{equation*}
    Similarly, the $k$-ary nested branching dimension of $\C$ is defined as
    \begin{equation*}
        \nbd_k(\C) \coloneqq \sup \{\rho_k(\C, K):K\in\CC\}.
    \end{equation*}
\end{definition}

The closure dimension and the nested branching dimension are incomparable, even for countable language classes, as shown in the lemma below. Its proof is given in Appendix~\ref{app:closure_nbd_incomparability}.

\begin{lemma}
\label{lem:closure_nbd_incomparability}
    There exist countable language classes $\C_1$ and $\C_2$ such that:
    \begin{itemize}
        \item $\cd(\C_1)=1$ and $\nbd(\C_1)=\infty$.
        \item $\cd(\C_2)=\infty$ and $\nbd(\C_2)=1$.
    \end{itemize}
\end{lemma}

We also consider whether the finiteness of the intersection of a nonempty, possibly infinite subfamily of $\C$ can be witnessed by a finite subfamily. The finite-intersection Helly number measures the smallest uniform bound on the size of such a witness. It can be seen as a variant of the standard notion of Helly number in combinatorics, defined as the minimum integer $k$ such that for every non-empty subfamily $\F\subseteq \C$ with empty intersection, there exists a non-empty $\G\subseteq \F$ of size at most $k$ whose intersection is empty. The term originates from the celebrated Helly’s theorem~\cite{helly1923}: any finite family of convex sets in $\mathbb R^d$ has a Helly number of at most $d+1$.

\begin{definition}[Finite-Intersection Helly Number]
    The finite-intersection Helly number of $\C$, denoted by $\h(\C)$, is the least integer $h\ge 1$ such that for every non-empty subfamily $\F\subseteq \C$ with $\left\lvert\bigcap \F\right\rvert < \infty$, there exists $\G\subseteq \F$ with $1\le \lvert \G\rvert\le h$ and $\lvert\bigcap\G\rvert <\infty$. If such an $h$ does not exist, then $\h(\C)=\infty$.
\end{definition}

The following lemma shows that the finite-intersection Helly property always holds for classes with finite nested branching dimension.

\begin{lemma}
\label{lem:nbd2h}
    For any language class $\C$, we have $\h(\CC) \le \nbd(\C) + 1$.
\end{lemma}
\begin{proof}
    The result trivially holds when $\nbd(\C) = \infty$. Now suppose $\nbd(\C) = d < \infty$ and $\h(\CC) > d + 1$. Then there exists a non-empty subfamily $\F\subseteq \CC$ whose total intersection is finite, but $\bigcap \G$ is infinite for every $\G\subseteq \F$ with $1\le \lvert\G\rvert\le d + 1$. Pick an arbitrary $F_1\in \F$. Construct a tree with node set $\V = \{(F_1,\dots,F_r):F_2,\dots,F_r\in\F, 1\le r\le d + 1\}$ such that a child extends its parent by exactly one coordinate. For every $v=(F_1,\dots,F_r)\in\V$, let $K_v = F_1\cap\cdots\cap F_r$. Since $r\le d+1$, $K_v$ is infinite and therefore $K_v\in \CC$. Moreover, $K_v\subseteq K_u$ whenever $u$ is a parent of $v$. Finally, for every internal node $u=(F_1,\dots,F_\ell)$ ($1\le \ell\le d$), we have that
    \begin{equation*}
        \bigcap_{\text{$v$ is a child of $u$}}K_v = \bigcap_{F_{\ell+1}\in\F}F_1\cap\cdots\cap F_{\ell+1} =\bigcap \F
    \end{equation*}
    is finite. Hence, this is a nested branching tree with depth $d+1$, contradicting that $\nbd(\C) = d$. Thus, $\h(\CC)\le d+1 = \nbd(\C) + 1$.
\end{proof}

The following lemma establishes that $\C$ and $\CC$ have the same finite-intersection Helly number.

\begin{lemma}
\label{lem:Helly_pass}
    For any language class $\C$, we have $\h(\C) = \h(\CC)$.
\end{lemma}
\begin{proof}
    Since $\C\subseteq \CC$, every $\F\subseteq \C$ is also a subfamily of $\CC$. Thus, we have $\h(\C) \le \h(\CC)$. If $\h(\C)=\infty$, equality follows, so assume $\h(\C)<\infty$. Conversely, pick a non-empty $\F\subseteq \CC$ with $\lvert \bigcap \F\rvert < \infty$. Let $\mathcal \C' = \{L\in \C: \exists F\in \F, F\subseteq L\}$. For every $F\in \F$, there exists $\G\subseteq \C$ such that $F = \bigcap \G$. Since $F\subseteq G$ for every $G\in\G$, we have $G\in \C'$ for every $G\in\G$. Therefore, $\bigcap \C'\subseteq \bigcap \G = F$. This further implies that $\bigcap\C'\subseteq \bigcap \F$ is also finite. Since $\C'\subseteq \C$, there exist $L_1,\cdots,L_h\in\C'$ with $h\le \h(\C)$ such that $L_1\cap\cdots\cap L_h$ is finite. For each $L_i$, there is some $F_i\in\F$ such that $F_i\subseteq L_i$. Consequently, $F_1\cap\cdots\cap F_h$ is also finite. Thus, $\h(\CC)\le\h(\C)$.
\end{proof}

\section{Upper Bound}

In this section, we present our algorithm for private uniform generation. Consider a non-empty language class $\C$ with $\cd(\C)<\infty$ and $\nbd(\C)<\infty$. If $\cd(\C)=-1$, then $\bigcap\C$ is an infinite subset of every $L\in\C$. The constant algorithm that outputs $\bigcap\C$ without reading the input is $(0,0)$-differentially private. Applying the randomized input-exclusion procedure of Theorem~\ref{thm:upper_exclude_inputs} immediately yields a $(0,\delta)$-differentially private uniform generator. We therefore assume $\cd(\C)\ge0$ in this section.

\subsection{Cores and Profiles}

For any $K\in\CC$, we define its core as the intersection of all sets in $\CC$ that are contained in $K$ and have the same rank as $K$.

\begin{definition}
    The core of $K\in\CC$ is defined as
    \begin{equation*}
        \core(K) \coloneqq \bigcap_{T\in\CC,\ T\subseteq K,\ \rho(T)=\rho(K)}T.
    \end{equation*}
\end{definition}

The following lemma shows that the core of $K$ is also in $\CC$.

\begin{lemma}
\label{lem:upper_core_infinite}
    For every $K\in\CC$, $\core(K)\in\CC$.
\end{lemma}
\begin{proof}
    Let $K\in\CC$ with $\rho(K)\le\nbd(\C)<\infty$, and let $\T=\{T\in\CC:T\subseteq K,\ \rho(T)=\rho(K)\}$. This family is non-empty because it contains $K$. Suppose that $\lvert\core(K)\rvert=\lvert\bigcap\T\rvert<\infty$. Beneath a new root labeled $K$, attach every nested branching tree of depth $\rho(K)$ rooted at a member of $\T$ (viewing trees as sets of finite labeled paths). Every $T\in\T$ contributes at least one such tree, so the intersection of the child labels is exactly $\bigcap\T$, which is finite. The resulting tree has depth $\rho(K)+1$, a contradiction. Hence, $\lvert\core(K)\rvert=\infty$. Set $\F=\{L\in\C:\exists T\in\T,\ T\subseteq L\}$. As each $T\in\T$ is an intersection of languages from $\C$, we have $\bigcap\F=\bigcap\T=\core(K)$, so $\core(K)\in\CC$.
\end{proof}

Since $\core(K)\in\CC$, its rank is well-defined. We have the following property for nested sets with the same rank.
\begin{lemma}
\label{lem:upper_core_nesting}
    Let $H,K\in\CC$ with $H\subseteq K$ and $\rho(H)=\rho(K)$. Then $\core(K)\subseteq\core(H)$. If additionally $\rho(\core(K))=\rho(H)=\rho(K)$, then $\core(H)=\core(K)$.
\end{lemma}
\begin{proof}
    Every $T\in\CC$ with $T\subseteq H$ and $\rho(T)=\rho(H)$ also satisfies $T\subseteq K$ and $\rho(T)=\rho(K)$. Thus, the family defining $\core(H)$ is contained in the family defining $\core(K)$, which gives $\core(K)\subseteq\core(H)$.

    Since $H$ itself belongs to the family defining $\core(K)$, we also have $\core(K)\subseteq H$. If $\rho(\core(K))=\rho(H)$, Lemma~\ref{lem:upper_core_infinite} implies that $\core(K)$ belongs to the family defining $\core(H)$. Consequently, $\core(H)\subseteq\core(K)$, and equality follows.
\end{proof}

Now consider the process of repeatedly applying $\core(\cdot)$. Since $\core(K)\subseteq K$, rank monotonicity gives two possibilities at each stage:
\begin{itemize}
    \item $\rho(\core(K))=\rho(K)$. In this case, $\core(K)$ is the smallest set in $\CC$ that is contained in $K$ and has the same rank as $K$. Applying Lemma~\ref{lem:upper_core_nesting} with $H=\core(K)$ gives $\core(\core(K))=\core(K)$. It can be seen as being at a stationary point of this operation.
    \item $\rho(\core(K))<\rho(K)$. In this case, applying $\core(\cdot)$ strictly reduces the rank. Since rank is a positive integer, this can occur at most $\rho(K) - 1$ times before reaching the first case.
\end{itemize}
Therefore, we can define the profile as the sequence of ranks in this process, and the iterated core as the final set.
\begin{definition}
    The profile of $K\in\CC$, denoted $\profile(K)$, is defined recursively by
    \begin{equation*}
        \profile(K)\coloneqq
        \begin{cases}
            (\rho(K)) &\text{if }\rho(\core(K))=\rho(K),\\
            (\rho(K))\frown\profile(\core(K)) &\text{if }\rho(\core(K))<\rho(K).
        \end{cases}
    \end{equation*}
    Here, $\frown$ denotes concatenation of sequences.
\end{definition}

\begin{definition}
    The iterated core of $K\in\CC$, denoted $\core^*(K)$, is defined recursively by
    \begin{equation*}
        \core^*(K)\coloneqq
        \begin{cases}
            \core(K) &\text{if }\rho(\core(K))=\rho(K),\\
            \core^*(\core(K)) &\text{if }\rho(\core(K))<\rho(K).
        \end{cases}
    \end{equation*}
\end{definition}

Since $\rho(K)\le \nbd(\C) < \infty$,  Lemma~\ref{lem:upper_core_infinite} gives $\core^*(K)\in\CC$.

Let $\Xi_d$ be the set of all strictly decreasing sequences with entries in $[d]$, including the empty sequence $()$. Then $\lvert\Xi_d\rvert=2^d$. We put a total order $\preceq_d$ on $\Xi_d$ recursively. For $d=1$, let $(1)\preceq_1()$. For $d\ge2$, consider two different sequences $\sigma_1,\sigma_2\in\Xi_d$. There are three cases:
\begin{itemize}
    \item Neither sequence starts with $d$. In this case, we preserve the order on $\Xi_{d-1}$, so $\sigma_1\preceq_d \sigma_2$ if and only if $\sigma_1\preceq_{d-1}\sigma_2$.
    \item Exactly one sequence starts with $d$. In this case, the sequence starting with $d$ precedes the other sequence.
    \item Both sequences start with $d$. If one of them is $(d)$, the other sequence precedes $(d)$. Otherwise, write $\sigma_i=(d)\frown\tau_i$ with $\tau_i\ne()$. We order their tails in the reverse of $\preceq_{d-1}$, so $\sigma_1\preceq_d \sigma_2$ if and only if $\tau_2\preceq_{d-1}\tau_1$.
\end{itemize}
Thus, the empty sequence is the largest element of $\Xi_d$, and every non-singleton sequence starting with $d$ precedes $(d)$.

The first case implies the following fact by induction.
\begin{fact}
\label{fact:upper_order_restriction}
    For every $d'\ge d\ge1$, the restriction of $\preceq_{d'}$ to $\Xi_d$ is $\preceq_d$. In particular, if $H,K\in\CC$ and $\profile(K)\preceq_d\profile(H)$ for some $d\ge\max(\rho(H),\rho(K))$, then $\profile(K)\preceq_{d'}\profile(H)$ for every $d'\ge d$.
\end{fact}

For every $K\in\CC$, we have $\profile(K)\in\Xi_{\nbd(\C)}\setminus\{()\}$. The following lemma establishes the monotonicity of profiles.

\begin{lemma}
\label{lem:upper_profile_monotonicity}
    If $H,K\in\CC$ and $H\subseteq K$, then $\profile(K)\preceq_{\nbd(\C)}\profile(H)$.
\end{lemma}
\begin{proof}
    By Fact~\ref{fact:upper_order_restriction}, it suffices to show $\profile(K)\preceq_{\rho(K)}\profile(H)$. We prove this by induction on $\rho(K)$. If $\rho(K)=1$, rank monotonicity gives $\rho(H)=1$. Both profiles are $(1)$, so the claim holds.

    Now suppose $\rho(K)\ge2$. If $\rho(H)<\rho(K)$, then $\profile(K)\preceq_{\rho(K)}\profile(H)$ because exactly the first sequence starts with $\rho(K)$. It remains to consider $\rho(H)=\rho(K)$. Lemma~\ref{lem:upper_core_nesting} gives $\core(K)\subseteq\core(H)$, and hence
    \begin{equation*}
        \rho(\core(K))\le\rho(\core(H))\le\rho(K).
    \end{equation*}
    There are three cases:
    \begin{itemize}
        \item $\rho(\core(K))=\rho(\core(H))=\rho(K)$. Then $\profile(H)=\profile(K)=(\rho(K))$.
        \item $\rho(\core(K))<\rho(\core(H))=\rho(K)$. Then $\profile(H)=(\rho(K))$, whereas $\profile(K)=(\rho(K))\frown\profile(\core(K))$ has a non-empty tail. By the third case in the definition of the order, $\profile(K)\preceq_{\rho(K)}\profile(H)$.
        \item $\rho(\core(K))\le\rho(\core(H))<\rho(K)$. The induction hypothesis applied to $\core(K)\subseteq\core(H)$, followed by Fact~\ref{fact:upper_order_restriction}, gives
        \begin{equation*}
            \profile(\core(H))\preceq_{\rho(K)-1}\profile(\core(K)).
        \end{equation*}
        Both tails are non-empty, so reversing their order after prefixing $\rho(K)$ gives $\profile(K)\preceq_{\rho(K)}\profile(H)$.
    \end{itemize}
    This completes the induction.
\end{proof}

The following lemma shows that nested sets with equal profiles have the same iterated cores. This provides the desired stability for achieving privacy.

\begin{lemma}
\label{lem:upper_equal_profiles}
    If $H,K\in\CC$, $H\subseteq K$, and $\profile(H)=\profile(K)$, then $\core^*(H)=\core^*(K)$.
\end{lemma}
\begin{proof}
    We prove the claim by induction on $\rho(K)$. Equality of profiles implies $\rho(H)=\rho(K)$. If the common profile is $(\rho(K))$, then $\rho(\core(K))=\rho(K)$, and Lemma~\ref{lem:upper_core_nesting} gives
    \begin{equation*}
        \core^*(H)=\core(H)=\core(K)=\core^*(K).
    \end{equation*}
    This includes the base case $\rho(K)=1$.

    Otherwise, $\rho(K)\ge2$ and the common profile is $(\rho(K))\frown\tau$ for some $\tau\in\Xi_{\rho(K)-1}\setminus\{()\}$. Lemma~\ref{lem:upper_core_nesting} gives $\core(K)\subseteq\core(H)$, and both cores have profile $\tau$ and rank less than $\rho(K)$. The induction hypothesis therefore yields
    \begin{equation*}
        \core^*(H)=\core^*(\core(H))=\core^*(\core(K))=\core^*(K).
    \end{equation*}
\end{proof}

\subsection{Robust Closures}

Recall that we assume $\cd(\C)\ge0$. It must be the case that $\h(\C)\ge 2$; otherwise, every intersection of languages from $\C$ would be infinite, implying $\cd(\C)=-1$. By Lemma~\ref{lem:Helly_pass} and Lemma~\ref{lem:nbd2h}, we have $\h(\CC) = \h(\C)\le\nbd(\C)+1<\infty$.

Let $S=(x_1,\dots,x_n)\in\X^n$ be arbitrary, with repetitions allowed. Let $m>\cd(\C)$ be a public integer whose value will be determined later. 

\begin{definition}[Robust Closure]
\label{def:robust_closure}
    For every integer $t$ with $0\le t < N(\C, m)$, where
\begin{equation*}
    N(\C, m)\coloneqq \left\lceil\frac{m-\cd(\C)}{\h(\C)}\right\rceil,
\end{equation*}
let $R_t(S) = \{T\subseteq \hat{S}:\lvert T\rvert = \lvert \hat S\rvert - t\text{ and }\exists L\in\C,\ T\subseteq L\}$ and define the robust closure
\begin{equation*}
    Q_t(S)\coloneqq
    \begin{cases}
        \varnothing &\text{if }\lvert\hat S\rvert-\h(\C)t\le\cd(\C),\\
        \bigcap_{T\in R_t(S)} \langle T\rangle_\C &\text{if }\lvert\hat S\rvert-\h(\C)t>\cd(\C)\text{ and }R_t(S)\neq \varnothing,\\
        \X &\text{otherwise.}
    \end{cases}
\end{equation*}
\end{definition}
For a fixed $t$, the above definition depends only on $\hat{S}$, the distinct support of $S$. In contrast, the range of $t$ depends only on the public threshold $m$ and is independent of $\hat{S}$.

The following lemma suggests that the second case in Definition~\ref{def:robust_closure} always gives an infinite set in $\CC$.
\begin{lemma}
\label{lem:upper_robust_infinite}
    If $\lvert\hat S\rvert-\h(\C)t>\cd(\C)$ and $R_t(S)\neq\varnothing$, then $Q_t(S)\in\CC$.
\end{lemma}
\begin{proof}
    Consider any $1\le k\le\h(\C)$ sets $T_1,\dots,T_k$ in $R_t(S)$. Since each of them omits exactly $t$ elements of $\hat S$,
    \begin{equation*}
        \left\lvert\bigcap_{i=1}^k T_i\right\rvert
        \ge\lvert\hat S\rvert-kt
        \ge\lvert\hat S\rvert-\h(\C)t
        >\cd(\C).
    \end{equation*}
    Their intersection is contained in a language from $\C$, and therefore its closure is infinite. By closure monotonicity, $\langle\bigcap_{i=1}^k T_i\rangle_\C\subseteq \langle T_j\rangle_\C$ for every $j\in[k]$ and hence
    \begin{equation*}
        \left\langle\bigcap_{i=1}^k T_i\right\rangle_\C
        \subseteq\bigcap_{j=1}^k\langle T_j\rangle_\C.
    \end{equation*}
    Thus, $\bigcap_{j=1}^k\langle T_j\rangle_\C$ is also infinite and belongs to $\CC$. Since $\h(\CC)=\h(\C)$, the finite-intersection Helly property implies that
    \begin{equation*}
        \bigcap_{T\in R_t(S)} \langle T\rangle_\C
    \end{equation*}
    is infinite and belongs to $\CC$.
\end{proof}

We prove a key property of robust closures, which implies that they form a nested sequence on any dataset and are interlacing on neighboring datasets.

\begin{lemma}
\label{lem:upper_interlacing}
    Let $S,S'\in\X^n$ differ in at most one entry. Whenever $0\le t<t+1<N(\C, m)$,
    \begin{equation*}
        Q_{t+1}(S)\subseteq Q_t(S').
    \end{equation*}
    In particular, $Q_{N(\C,m)-1}(S)\subseteq\dots\subseteq Q_0(S)$ for any dataset $S$.  For neighboring datasets $S$ and $S'$, we have
    \begin{equation*}
        Q_{t+1}(S)\subseteq Q_t(S'),
        \qquad
        Q_{t+1}(S')\subseteq Q_t(S).
    \end{equation*}
\end{lemma}
\begin{proof}
    The claim is immediate if $Q_{t+1}(S)=\varnothing$ or $Q_t(S')=\X$. If $Q_t(S')=\varnothing$, Lemma~\ref{lem:upper_robust_infinite} and the definition imply $\lvert\hat S'\rvert-\h(\C)t\le\cd(\C)$. Since $\lvert\hat S\rvert\le\lvert\hat S'\rvert+1$,
    \begin{equation*}
    \begin{aligned}
        \lvert\hat S\rvert-\h(\C)(t+1)
        &\le\lvert\hat S'\rvert+1-\h(\C)t-\h(\C)\\
        &\le\cd(\C)+1-\h(\C)\\
        &\le\cd(\C).
    \end{aligned}
    \end{equation*}
    Hence $Q_{t+1}(S)=\varnothing$ in this case too.

    It remains to consider $Q_t(S')\notin\{\varnothing,\X\}$ and $Q_{t+1}(S)\ne\varnothing$. In this case $R_t(S')$ is non-empty. Take any set $T'\in R_t(S')$. Since neighboring supports satisfy $\lvert\hat S\setminus\hat S'\rvert\le1$,
    \begin{equation*}
        \lvert\hat S\setminus T'\rvert
        \le\lvert\hat S\setminus\hat S'\rvert
           +\lvert\hat S'\setminus T'\rvert
        \le t+1.
    \end{equation*}
    The assumption $Q_{t+1}(S)\ne\varnothing$ also gives $\lvert\hat S\rvert-\h(\C)(t+1)>\cd(\C)\ge0$. Hence, it follows that 
    \begin{equation*}
        \lvert\hat{S} \cap T'\rvert = \lvert\hat{S}\rvert - \lvert\hat{S}\setminus T'\rvert \ge \lvert \hat S\rvert - t - 1 > 0.
    \end{equation*}We can therefore choose a non-empty set $T\subseteq\hat S\cap T'$ with $\lvert T\rvert=\lvert\hat S\rvert-t-1$. Since $T'$ is contained in a language from $\C$, so is $T$. Thus $T\in R_{t+1}(S)$, and closure monotonicity gives
    \begin{equation*}
        Q_{t+1}(S)\subseteq\langle T\rangle_\C
        \subseteq\langle T'\rangle_\C.
    \end{equation*}
    Intersecting over all choices of $T'$ proves the desired inclusion. Taking $S'=S$ gives nestedness, and exchanging $S,S'$ gives the second interlacing inclusion.
\end{proof}

\subsection{Stability of Profile Blocks}

To treat every possible value of $Q_t(S)$, let $\overline{\profile}$ and $\overline{\core^*}$ agree with $\profile$ and $\core^*$ on $\CC$, and set
\begin{equation*}
    \overline{\profile}(\varnothing)=(),
    \qquad \overline{\core^*}(\varnothing)=\X.
\end{equation*}
If $\X\notin\CC$, also set
\begin{equation*}
    \overline{\profile}(\X)=(\nbd(\C)+1),
    \qquad \overline{\core^*}(\X)=\X.
\end{equation*}
If $\X\in\CC$, the barred maps retain their original values there.

For the remainder of the section, write
\begin{equation*}
    \Xi\coloneqq
    \Xi_{\nbd(\C)}\cup\{(\nbd(\C)+1)\},
\end{equation*}
and let $\preceq$ denote the restriction of $\preceq_{\nbd(\C)+1}$ to $\Xi$. Then
\begin{equation*}
    \lvert\Xi\rvert=2^{\nbd(\C)}+1.
\end{equation*}
The profile $(\nbd(\C)+1)$ is the least element of $\Xi$, and $()$ is its greatest element. By Fact~\ref{fact:upper_order_restriction}, this order agrees with $\preceq_{\nbd(\C)}$ on the profiles of members of $\CC$. Also, monotonicity and equality naturally extend to augmented profiles and iterated cores.

\begin{lemma}
\label{lem:monotonicity_equality_extend}
    If $H, K\in\CC\cup\{\varnothing,\X\}$ and $H\subseteq K$, then $\overline{\profile}(K)\preceq \overline{\profile}(H)$. If additionally $\overline{\profile}(H) = \overline{\profile}(K)$, then $\overline{\core^*}(H) = \overline{\core^*}(K)$.
\end{lemma}
\begin{proof}
    On $\CC$, this follows from Lemmas~\ref{lem:upper_profile_monotonicity} and~\ref{lem:upper_equal_profiles}. The empty profile occurs only at $\varnothing$ and is greatest. If $\X\notin\CC$, its profile $(\nbd(\C)+1)$ occurs only at $\X$ and is least; otherwise comparisons involving $\X$ are already covered. Thus monotonicity holds in every remaining case, and equal profiles there force $H=K$.
\end{proof}

Lemma~\ref{lem:upper_interlacing} gives $Q_{N(\C,m)-1}(S)\subseteq\dots\subseteq Q_0(S)$. Then by Lemma~\ref{lem:monotonicity_equality_extend}, the profile sequence is nondecreasing:
\begin{equation*}
    \overline{\profile}(Q_0(S))\preceq\dots\preceq\overline{\profile}(Q_{N(\C,m)-1}(S)).
\end{equation*} Hence, those $Q_t(S)$ with the same profile form a contiguous block. This motivates the following definition.

\begin{definition}
\label{def:length_common_core}
    For $\sigma\in\Xi$, define its block length by
\begin{equation*}
    \lambda_S(\sigma)\coloneqq\left\lvert\left\{0\le t<N(\C, m): \overline{\profile}(Q_t(S))=\sigma\right\}\right\rvert.
\end{equation*}
    If $\lambda_S(\sigma) > 0$, then there is some $t$ with $\overline{\profile}(Q_t(S))=\sigma$. We can define
    \begin{equation*}
        \gamma_S(\sigma) \coloneqq \overline{\core^*}(Q_{t_0}(S)),
    \end{equation*}
    where $t_0$ is the least index with $\overline{\profile}(Q_{t_0}(S)) = \sigma$.
\end{definition}

In the above definition, $\gamma_S(\sigma)$ is defined as the iterated core of $Q_{t_0}(S)$. Indeed, it is equal to every $\overline{\core^*}(Q_{t}(S))$ with $\overline{\profile}(Q_t(S))=\sigma$, as shown in the following fact. Thus, it represents the common iterated core of a contiguous block with the same profile.

\begin{fact}
    $\gamma_S(\sigma) = \overline{\core^*}(Q_t(S))$ for every $t$ with $\overline{\profile}(Q_t(S)) = \sigma$.
\end{fact}
\begin{proof}
    Since $t_0$ is the least index with $\overline{\profile}(Q_{t_0}(S)) = \sigma$, Lemma~\ref{lem:upper_interlacing} gives $Q_t(S)\subseteq Q_{t_0}(S)$. The conclusion follows from Lemma~\ref{lem:monotonicity_equality_extend} by setting $H = Q_t(S)$ and $K = Q_{t_0}(S)$.
\end{proof}

Next, we bound the sensitivity of $\lambda_S(\sigma)$ and prove the stability of $\gamma_S(\sigma)$.
\begin{lemma}
\label{lem:upper_block_stability}
    For neighboring datasets $S,S'\in\X^n$ and every $\sigma\in\Xi$,
    \begin{equation*}
        \lvert\lambda_S(\sigma)-\lambda_{S'}(\sigma)\rvert\le2.
    \end{equation*}
    Moreover, for every $\sigma$ with $\lambda_S(\sigma)>2$, we have $\lambda_{S'}(\sigma) > 0$ and $\gamma_S(\sigma) = \gamma_{S'}(\sigma)$.
\end{lemma}
\begin{proof}
    Lemma~\ref{lem:upper_interlacing} and Lemma~\ref{lem:monotonicity_equality_extend} give
    \begin{equation*}
        \overline{\profile}(Q_t(S'))\preceq\overline{\profile}(Q_{t+1}(S)),
        \qquad
        \overline{\profile}(Q_t(S))\preceq\overline{\profile}(Q_{t+1}(S')).
    \end{equation*}

    For $\sigma\in\Xi$, define the prefix sum $\Lambda_S(\sigma)=\sum_{\tau\in\Xi,\,\tau\preceq \sigma}\lambda_S(\tau)$. The $t$'s with $\overline{\profile}(Q_t(S))\preceq \sigma$ form the first $\Lambda_S(\sigma)$ indices. If $\Lambda_S(\sigma)\ge2$, the first inequality above gives
    \begin{equation*}
        \overline{\profile}(Q_{\Lambda_S(\sigma)-2}(S')) \preceq\overline{\profile}(Q_{\Lambda_S(\sigma)-1}(S)) \preceq \sigma.
    \end{equation*}
    Thus, $\Lambda_{S'}(\sigma)\ge \Lambda_S(\sigma)-1$. This is trivial when $\Lambda_S(\sigma)\le1$. By symmetry, $\lvert\Lambda_S(\sigma)-\Lambda_{S'}(\sigma)\rvert\le1$. If $\sigma$ is the first element of $\Xi$, then $\lambda_S(\sigma)=\Lambda_S(\sigma)$, so the desired bound follows. Otherwise, let $\sigma^-$ immediately precede $\sigma$. Then
    \begin{align*}
        \lvert\lambda_S(\sigma) - \lambda_{S'}(\sigma)\rvert &= \lvert\Lambda_S(\sigma) - \Lambda_S(\sigma^-) - \Lambda_{S'}(\sigma) + \Lambda_{S'}(\sigma^-)\rvert \\
        &\le \lvert\Lambda_S(\sigma) - \Lambda_{S'}(\sigma)\rvert + \lvert \Lambda_S(\sigma^-) -\Lambda_{S'}(\sigma^-)\rvert \\
        &\le 2.
    \end{align*}

    Now suppose $\lambda_S(\sigma)>2$. The preceding result gives $\lambda_{S'}(\sigma)>0$. Choose consecutive $t$ and $t+1$ in the $\sigma$-block for $S$, i.e., $\overline{\profile}(Q_t(S))=\overline{\profile}(Q_{t+1}(S))=\sigma$. Therefore,
    \begin{equation*}
        \overline{\profile}(Q_t(S'))\preceq\overline{\profile}(Q_{t+1}(S)) = \sigma
    \end{equation*}
    and
    \begin{equation*}
        \sigma = \overline{\profile}(Q_t(S))\preceq\overline{\profile}(Q_{t+1}(S')).
    \end{equation*}
    Because the profile sequence for $S'$ is nondecreasing and $\lambda_{S'}(\sigma) > 0$, at least one of these two comparisons is an equality. In either case Lemma~\ref{lem:monotonicity_equality_extend} gives $\gamma_S(\sigma) = \gamma_{S'}(\sigma)$.
\end{proof}

\subsection{Private Selection}

On input $S\in\X^n$, our algorithm chooses an occurring profile $\sigma$ (i.e., $\lambda_S(\sigma)>0$) with probability
\begin{equation*}
    \frac{\exp\left(\varepsilon\lambda_S(\sigma)/8\right)}{\sum_{\tau\in\Xi,            \lambda_S(\tau)>0}\exp\left(\varepsilon\lambda_S(\tau)/8\right)}.
\end{equation*}
It then returns $\gamma_S(\sigma)$ (Definition~\ref{def:length_common_core}). 
\begin{theorem}
\label{thm:upper_private_selection}
    Let $\C$ be a language class with $0\le\cd(\C)<\infty$ and $\nbd(\C)<\infty$, and let $0<\varepsilon,\delta<1$. Fix a public integer $m$ satisfying
    \begin{equation*}
        m\ge\cd(\C)+\frac{8\h(\C)(2^{\nbd(\C)}+1)}{\varepsilon}\max\left\{\ln\frac{2(2^{\nbd(\C)}+1)}{\varepsilon},\,\ln\frac{2^{\nbd(\C)}+1}{\delta}+\frac{\varepsilon}{4}\right\}.
    \end{equation*}
    Then, for every input length $n$, there exists an $(\varepsilon,\delta)$-differentially private algorithm $\A$ that outputs an infinite subset of $\X$ almost surely on every input $S\in\X^n$ and, when $\hat S$ is contained in a language from $\C$ and $\lvert\hat S\rvert\ge m$,
    \begin{equation*}
        \Pr\left[
            \lvert\A(S)\rvert=\infty
            \text{ and }
            \A(S)\subseteq\langle\hat S\rangle_\C
        \right]=1.
    \end{equation*}
    In particular, a sufficient threshold on the number of distinct examples is
    \begin{equation*}
        m=\cd(\C)+O\!\left(
            \frac{\h(\C)2^{\nbd(\C)}}{\varepsilon}
            \left(\nbd(\C)+\log\frac1\varepsilon+\log\frac1\delta\right)
        \right).
    \end{equation*}
\end{theorem}
\begin{proof}
    Write $d=\nbd(\C)$, $h=\h(\C)$, $c=\cd(\C)$, $M=\lvert\Xi\rvert$, and $N=\lceil(m-c)/h\rceil = N(\C,m)$. For any dataset $S$, let $P_S=\{\sigma\in\Xi:\lambda_S(\sigma)>0\}$ and
    \begin{equation*}
        Z_S=\sum_{\sigma\in P_S}e^{\varepsilon\lambda_S(\sigma)/8}.
    \end{equation*}
    Construct an algorithm $\A$ that first samples $\sigma$ from $P_S$ with probability $\exp(\varepsilon\lambda_S(\sigma) / 8)/Z_S$, and then outputs $\gamma_S(\sigma)$.
    Definition~\ref{def:length_common_core} ensures that $\gamma_S(\sigma)$ is well-defined because $\lambda_S(\sigma)>0$. Since $m > c$, we have $N \ge 1$. This implies that $\lambda_S(\overline{\profile}(Q_0(S))) > 0$ and hence $P_S$ is non-empty. Thus, the sampling process is valid. By Lemma~\ref{lem:upper_robust_infinite} and our definition of $\overline{\core^*}$ ($\overline{\core^*}(\varnothing) = \overline{\core^*}(\X) = \X$), $\A$ produces an infinite subset of $\X$ almost surely.
    
    We first prove the privacy guarantee. The profile sequence of length $N$ is partitioned into at most $M$ profile blocks, so some block has length at least $N/M$. Hence
    \begin{equation*}
        Z_S\ge e^{\varepsilon N/(8M)}
        \ge e^{\varepsilon(m-c)/(8hM)}.
    \end{equation*}
    Recalling that $M=2^d+1$, our choice of $m$ ensures that
    \begin{equation*}
        \frac{\varepsilon(m-c)}{8hM}
        \ge
        \max\left\{
            \ln\frac{2M}{\varepsilon},
            \,\ln\frac{M}{\delta}+\frac{\varepsilon}{4}
        \right\}.
    \end{equation*}
    Combining the preceding lower bound on $Z_S$ with the first term in this maximum gives
    \begin{equation*}
        Z_S\ge\exp\left(\ln\frac{2M}{\varepsilon}\right)
        =\frac{2M}{\varepsilon}.
    \end{equation*}
    Similarly, the second term gives
    \begin{equation*}
        Z_S\ge\exp\left(\ln\frac{M}{\delta}+\frac{\varepsilon}{4}\right)
        =\frac{Me^{\varepsilon/4}}{\delta}.
    \end{equation*}

    Let $S,S'$ be neighboring datasets. By Lemma~\ref{lem:upper_block_stability}, we have $\lambda_{S'}(\sigma)\le\lambda_S(\sigma)+2$ for every $\sigma\in\Xi$. Consequently,
    \begin{equation*}
        e^{\varepsilon\lambda_{S'}(\sigma)/8}
        \le e^{\varepsilon/4}e^{\varepsilon\lambda_S(\sigma)/8}.
    \end{equation*}
    For $\sigma\in P_{S'}\setminus P_S$, we have $\lambda_S(\sigma)=0$, so the preceding inequality gives
    \begin{equation*}
        e^{\varepsilon\lambda_{S'}(\sigma)/8}\le e^{\varepsilon/4}.
    \end{equation*}
    Since there are at most $M$ such profiles, their total contribution to $Z_{S'}$ is at most $Me^{\varepsilon/4}$. Thus,
    \begin{align*}
        Z_{S'}
        &=\sum_{\sigma\in P_S\cap P_{S'}}
        e^{\varepsilon\lambda_{S'}(\sigma)/8}
        +\sum_{\sigma\in P_{S'}\setminus P_S}
        e^{\varepsilon\lambda_{S'}(\sigma)/8}\\
        &\le e^{\varepsilon/4}
        \sum_{\sigma\in P_S\cap P_{S'}}
        e^{\varepsilon\lambda_S(\sigma)/8}
        +Me^{\varepsilon/4}\\
        &\le e^{\varepsilon/4}(Z_S+M).
    \end{align*}
    Since $Z_S\ge2M/\varepsilon$, it follows that
    \begin{equation*}
        \frac{Z_{S'}}{Z_S}
        \le e^{\varepsilon/4}\left(1+\frac{M}{Z_S}\right)
        \le e^{\varepsilon/4}\left(1+\frac{\varepsilon}{2}\right)
        \le e^{3\varepsilon/4}.
    \end{equation*}
    For $\sigma\in P_S$, let
    \begin{equation*}
        q_S(\sigma)=\frac{e^{\varepsilon\lambda_S(\sigma)/8}}{Z_S}
    \end{equation*}
    be the probability of selecting $\sigma$ on input $S$, and define $q_{S'}$ analogously. Let
    \begin{equation*}
        G=\{\sigma\in P_S:\lambda_S(\sigma)>2\}.
    \end{equation*}
    By Lemma~\ref{lem:upper_block_stability}, every $\sigma\in G$ belongs to $P_{S'}$ and satisfies $\gamma_S(\sigma)=\gamma_{S'}(\sigma)$ and $\lambda_S(\sigma)-\lambda_{S'}(\sigma)\le2$.
    Combining this with the preceding bound on $Z_{S'}/Z_S$, we obtain
    \begin{align*}
        \frac{q_S(\sigma)}{q_{S'}(\sigma)}
        &=e^{\varepsilon(\lambda_S(\sigma)-\lambda_{S'}(\sigma))/8}
        \frac{Z_{S'}}{Z_S}\\
        &\le e^{\varepsilon/4}e^{3\varepsilon/4}
        =e^\varepsilon.
    \end{align*}
    For every $\sigma\in P_S\setminus G$, we have $\lambda_S(\sigma)\le2$. Since there are at most $M$ such profiles and $Z_S\ge Me^{\varepsilon/4}/\delta$,
    \begin{equation*}
        \sum_{\sigma\in P_S\setminus G}q_S(\sigma)
        =\frac{1}{Z_S}\sum_{\sigma\in P_S\setminus G}
        e^{\varepsilon\lambda_S(\sigma)/8}
        \le\frac{Me^{\varepsilon/4}}{Z_S}
        \le\delta.
    \end{equation*}
    Now let $O$ be any measurable output event. Since $\gamma_S(\sigma)=\gamma_{S'}(\sigma)$ for $\sigma\in G$,
    \begin{align*}
        \Pr[\A(S)\in O]
        &=\sum_{\sigma\in G:\,\gamma_S(\sigma)\in O}q_S(\sigma)
        +\sum_{\sigma\in P_S\setminus G:\,\gamma_S(\sigma)\in O}q_S(\sigma)\\
        &\le e^\varepsilon
        \sum_{\sigma\in G:\,\gamma_{S'}(\sigma)\in O}q_{S'}(\sigma)
        +\delta\\
        &\le e^\varepsilon\Pr[\A(S')\in O]+\delta.
    \end{align*}

    For utility, suppose $\hat S\subseteq L$ for some $L\in\C$ and $\lvert\hat S\rvert\ge m$. Every $0\le t<N$ satisfies $t<(m-c)/h$, so
    \begin{equation*}
        \lvert\hat S\rvert-ht\ge m-ht>c\ge0.
    \end{equation*}
    In particular, $t<\lvert\hat S\rvert$. Since $\hat S\subseteq L$, every subset of $\hat S$ of size $\lvert\hat S\rvert-t$ is contained in $L$. Thus $R_t(S)\ne\varnothing$, and Lemma~\ref{lem:upper_robust_infinite} gives $Q_t(S)\in\CC$.

    Since $R_0(S)=\{\hat S\}$, we have $Q_0(S)=\langle\hat S\rangle_\C$. Lemma~\ref{lem:upper_interlacing} therefore gives, for every $0\le t<N$,
    \begin{equation*}
        Q_t(S)\subseteq Q_0(S)=\langle\hat S\rangle_\C.
    \end{equation*}

    Every profile $\sigma$ selected by the algorithm occurs at some $t$. Since $Q_t(S)\in\CC$, its iterated core also belongs to $\CC$ and is therefore infinite. The corresponding output satisfies
    \begin{equation*}
        \gamma_S(\sigma)
        =\core^*(Q_t(S))
        \subseteq Q_t(S)
        \subseteq\langle\hat S\rangle_\C
        \subseteq L.
    \end{equation*}
    Thus every possible output is an infinite subset of $L$.
\end{proof}

\subsection{Excluding Input Examples}

The algorithm analyzed in Theorem~\ref{thm:upper_private_selection} may return a set containing input examples. The following theorem shows how to obtain an infinite subset of this set that excludes all input examples, with an arbitrarily small additive increase in $\delta$.

\begin{theorem}
\label{thm:upper_exclude_inputs}
    Let $\A$ be an $(\varepsilon,\delta)$-differentially private algorithm that produces an infinite subset of $\X$ almost surely on every input. For every $\eta>0$, there is an $(\varepsilon,\delta+\eta)$-differentially private algorithm that first samples $K=\A(S)$ and then outputs an infinite subset of $K\setminus\hat S$ almost surely.
\end{theorem}
\begin{proof}
    Fix an enumeration of $\X$. For every infinite $K$, let $k_1,k_2,\dots$ denote its elements in this order. Let $b=\lceil 2e^\varepsilon/\eta\rceil$. After obtaining $K=\A(S)$, choose $J$ uniformly from $[b]$, independently of $K$, and output
    \begin{equation*}
        \A_\eta(S)=\{k_{J+rb}:r=0,1,\ldots\}\setminus\hat S.
    \end{equation*}
    This set is infinite and has the required inclusion because only finitely many elements are removed.

    For neighboring $S,S'$ and any fixed $K$, consider the same $J$ in both constructions. Their outputs can differ only if the selected residue subset contains a point belonging to exactly one of $\hat S,\hat S'$. There are at most two such points, each selected with probability at most $1/b$. Hence, for any measurable output event $O$, if
    \begin{equation*}
        f_S(K)=\Pr_J\!\left[\{k_{J+rb}:r=0,1,\ldots\}\setminus\hat S\in O\right],
    \end{equation*}
    then $\lvert f_S(K)-f_{S'}(K)\rvert\le2/b$. Differential privacy bounds expectations of $[0,1]$-valued functions by integrating their level sets. Therefore
    \begin{align*}
        \Pr[\A_\eta(S)\in O]
        &=\mathbb E_{K\sim\A(S)}f_S(K)\\
        &\le e^\varepsilon\mathbb E_{K\sim\A(S')}f_S(K)+\delta\\
        &\le e^\varepsilon\mathbb E_{K\sim\A(S')}f_{S'}(K)+\delta+2e^\varepsilon/b\\
        &\le e^\varepsilon\Pr[\A_\eta(S')\in O]+\delta+\eta.
    \end{align*}
\end{proof}

Combining this theorem with Theorem~\ref{thm:upper_private_selection}, using $\delta/2$ there and $\eta=\delta/2$ here, proves our main upper bound. 

\begin{algorithm}[!ht]
\label{alg:gen}
\DontPrintSemicolon
    \KwInput{Language class $\C$, dataset $S$, privacy parameters $(\varepsilon,\delta)$}
    \KwOutput{An infinite set $R\subseteq \X$}
    $m\gets \left\lceil\cd(\C)+\frac{8\h(\C)(2^{\nbd(\C)}+1)}{\varepsilon}\max\left\{\ln\frac{2(2^{\nbd(\C)}+1)}{\varepsilon},\,\ln\frac{2(2^{\nbd(\C)}+1)}{\delta}+\frac{\varepsilon}{4}\right\}\right\rceil$\;
    Construct $Q_0(S),\dots,Q_{N(\C,m)-1}(S)$ (Definition~\ref{def:robust_closure})\;
    Sample $\sigma$ from $P_S=\{\sigma\in\Xi:\lambda_S(\sigma)>0\}$ with probability $\frac{\exp\left(\varepsilon\lambda_S(\sigma)/8\right)}{\sum_{\tau\in P_S}\exp\left(\varepsilon\lambda_S(\tau)/8\right)}$ and set $K\gets \gamma_S(\sigma)$ (Definition~\ref{def:length_common_core})\;
    Order elements in $K$ according to a fixed enumeration of $\X$: $k_1,k_2,\dots$\;
    $R\gets \{k_{J+r\lceil4e^\varepsilon / \delta\rceil}:r=0,1,\ldots\}\setminus\hat S$, where $J$ is sampled uniformly from $[\lceil4e^\varepsilon / \delta\rceil]$\;
    \Return{$R$}
\caption{Private Uniform Language Generation}
\end{algorithm}

\section{Lower Bound}
In this section, we prove our lower bound. We will prove a lower bound that includes the finite-intersection Helly number, as stated below. It is stronger than our result stated in the introduction.

\begin{theorem}
\label{thm:lower_full}
    Let $n\ge 1$, $0\le \varepsilon \le 1$, and $0<\delta\le 1/(12n)$. If $\A$ is an $(\varepsilon,\delta)$-differentially private uniform generation algorithm for language class $\C$ with sample size $n$, then
    \begin{equation*}
        n\ge \max(\cd(\C) + 1, \h(\C), \Omega(\log^*\nbd(\C))).
    \end{equation*}
\end{theorem}

We first pass from the utility guarantee for individual languages in Definition~\ref{def:uniform_language_generation} to a guarantee for their intersection. The countability of $\X$ is useful here. 

\begin{lemma}
\label{lem:lower_closure}
    If $\A$ is a uniform generation algorithm for $\C$ with sample size $n$ and $S$ is an input sequence consisting of $n$ distinct elements from some $L\in\C$, then $\A(S)$ is almost surely an infinite subset of $\langle\hat S\rangle_\C$.
\end{lemma}
\begin{proof}
    Infiniteness follows directly from the definition of uniform generation. For each $x\notin\langle\hat S\rangle_\C$, some $L_x\in\C$ contains $\hat S$ but excludes $x$. Thus $\Pr[x\in\A(S)]=0$. Since $\X$ is countable, 
    \begin{equation*}
        \Pr[\A(S)\not\subseteq\langle\hat S\rangle_\C]
        \le\sum_{x\notin\langle\hat S\rangle_\C}\Pr[x\in\A(S)]
        =0.
    \end{equation*}
    Therefore, $\Pr[\A(S)\subseteq\langle\hat S\rangle_\C]=1$.
\end{proof}

We shall use two consequences of the definition of $\CC$. If $K\in\CC$ contains $\hat S$, then $\langle\hat S\rangle_\C\subseteq K$. Also, if a non-empty subfamily $\F\subseteq\CC$ has an infinite intersection, that intersection belongs to $\CC$: it equals the intersection of $\{L\in\C:\exists K\in\F,\ K\subseteq L\}$.

The $\cd(\C) + 1$ lower bound holds for non-private generators and was proved by Raman, Li, and Tewari~\cite{raman2025generation}. While they focused on deterministic algorithms, their proof naturally extends to randomized algorithms by Lemma~\ref{lem:lower_closure}.
\begin{lemma}[\cite{raman2025generation}]
\label{lem:lower_closure_dimension}
    If $\A$ is a uniform generation algorithm for language class $\C$ with sample size $n$, then $n > \cd(\C)$.
\end{lemma}
\begin{proof}
    Suppose $n\le \cd(\C)$. By the definition of the closure dimension, there is a finite set $T$ contained in some $L\in\C$ such that $\lvert T\rvert\ge n$ and $\lvert \langle T\rangle_\C\rvert <\infty$. Construct an input sequence $S$ from $n$ arbitrary distinct elements of $T$. Then $\hat{S}\subseteq T$. By Lemma~\ref{lem:lower_closure}, $\A(S)$ is almost surely an infinite subset of $\langle\hat{S}\rangle_\C\subseteq\langle T\rangle_\C$, contradicting that $\langle T\rangle_\C$ is finite.
\end{proof}

\subsection{Helly Obstruction}

The second obstruction is expressed through the finite-intersection Helly number defined in Section~\ref{sec:dimension}. The next lemma follows from the fingerprinting results of~\cite{bun2015differentially}. We include a proof in Appendix~\ref{app:interior_point_proofs} for completeness.

\begin{lemma}[\cite{bun2015differentially}]
\label{lem:lower_sharp_interior_point}
    Fix $n\ge1$, $0\le\varepsilon<\infty$, and $0\le\delta<1/n$. There are $N\ge2$ and $\eta>0$ such that, for every $(\varepsilon,\delta)$-differentially private algorithm $\M:[N]^n\to[N]$, some $y\in[N]^n$ satisfies
    \begin{equation*}
        \Pr\left[\min_{i\in[n]}y_i\le \M(y)\le\max_{i\in[n]}y_i\right]<1-\eta.
    \end{equation*}
\end{lemma}

\begin{lemma}
\label{lem:lower_helly}
    If $\A$ is an $(\varepsilon, \delta)$-differentially private uniform generation algorithm for $\C$ with sample size $n$ and $\delta < 1/n$, then $n\ge \h(\C)$.
\end{lemma}
\begin{proof}
    By Lemma~\ref{lem:Helly_pass}, it suffices to prove $\h(\CC)\le n$. Suppose instead that $\h(\CC)>n$. Then there exists a non-empty family $\F\subseteq\CC$ whose total intersection is finite, while every non-empty subfamily of at most $n$ members has infinite intersection. There are exactly two cases:
    \begin{itemize}
        \item Some finite subfamily of $\F$ has finite intersection.
        \item Every finite subfamily of $\F$ has infinite intersection.
    \end{itemize}

    In the first case, choose a finite subfamily $\{K_1,\dots,K_r\}\subseteq\F$ of minimum size whose intersection is finite. Then $r\ge n+1$, and minimality ensures that omitting any one of these sets leaves an infinite intersection. For each $i\in[r]$, define
    \begin{equation*}
        P_i=\left(\bigcap_{j\in[r]\setminus\{i\}}K_j\right)
        \setminus\bigcap_{j=1}^r K_j.
    \end{equation*}
    Each $P_i$ is infinite because it is obtained by removing a finite set from an infinite set. Every point in $P_i$ belongs to all $K_j$ with $j\ne i$ and lies outside $K_i$. Consequently, the sets $P_i$ are disjoint, and $P_i\subseteq K_r$ for every $i\in[r-1]$.

    Suppose $n\ge2$. Choose distinct points $a_{i,t}\in P_i$ for $i\in[r-1]$ and $t\in[n]$. Encode each $y\in[r-1]^n$ by $S_y=(a_{y_1,1},\dots,a_{y_n,n})$. The entries of $S_y$ are distinct and belong to $K_r$. By Lemma~\ref{lem:lower_closure}, $\A(S_y)$ is infinite and contained in $K_r$ almost surely. If $i$ does not occur in $y$, then every encoded point belongs to $K_i$, so the same lemma gives $\A(S_y)\subseteq K_i$ almost surely.

    For a set $Z\subseteq\X$, define
    \begin{equation*}
        V(Z)=\{i\in[r-1]:Z\not\subseteq K_i\}.
    \end{equation*}
    The preceding containment property implies that $V(\A(S_y))\subseteq\{y_1,\dots,y_n\}$ almost surely. Moreover, $V(\A(S_y))$ is non-empty almost surely: otherwise, the infinite output would be contained in the finite intersection $\bigcap_{j=1}^r K_j$.

    Let $\M(y)$ be the least element of $V(\A(S_y))$, using $1$ if this set is empty. Thus $\M(y)\in\{y_1,\dots,y_n\}$ almost surely. Changing one coordinate of $y$ changes only one entry of $S_y$, and $\M$ is a post-processing of $\A$. Hence $\M$ is $(\varepsilon,\delta)$-differentially private.

    Let $y_0=(1,\dots,n)$. For each $i\in[n]$, obtain $y_i$ by replacing the $i$-th coordinate of $y_0$ with any value in $[n]\setminus\{i\}$. Since $i$ does not occur in $y_i$, we have $\Pr[\M(y_i)=i]=0$. Privacy therefore gives
    \begin{equation*}
        \Pr[\M(y_0)=i]
        \le e^\varepsilon\Pr[\M(y_i)=i]+\delta
        =\delta.
    \end{equation*}
    Summing over $i\in[n]$, we obtain
    \begin{equation*}
        1=\sum_{i=1}^n\Pr[\M(y_0)=i]
        \le n\delta<1,
    \end{equation*}
    a contradiction.

    If $n=1$, choose $x\in P_1$ and $x'\in P_r$. Since $x\in\bigcap_{j=2}^rK_j$, Lemma~\ref{lem:lower_closure} implies that $\A((x))$ is infinite and contained in this intersection almost surely. It therefore cannot also be contained in $K_1$, because $\bigcap_{j=1}^rK_j$ is finite. Since $x'\in K_1$, we also have $\A((x'))\subseteq K_1$ almost surely. Privacy therefore gives
    \begin{equation*}
        1=\Pr[\A((x))\not\subseteq K_1]
        \le e^\varepsilon\Pr[\A((x'))\not\subseteq K_1]+\delta
        =\delta<1,
    \end{equation*}
    a contradiction.

    Now consider the second case. Fix $N\ge1$ and $\eta>0$. Since $\X$ is countable, let $z_1,z_2,\dots$ enumerate $\X\setminus\bigcap\F$. We recursively construct decreasing sets $B_0,\dots,B_N\in\CC$, each a finite intersection of members of $\F$, and distinct points $a_{i,t}$ for $i\in[N]$ and $t\in[n]$, satisfying
    \begin{equation*}
        a_{i,t}\in B_{i-1}\setminus\bigcap\F.
    \end{equation*}

    Choose any $B_0\in\F$. At stage $i$, the set $B_{i-1}$ is infinite and $\bigcap\F$ is finite, so we can choose $n$ previously unused points $a_{i,1},\dots,a_{i,n}$ from $B_{i-1}\setminus\bigcap\F$. For each $y\in[i]^n$, define $S_y=(a_{y_1,1},\dots,a_{y_n,n})$. Its entries are distinct and belong to $B_0$, so Lemma~\ref{lem:lower_closure} implies that $\A(S_y)$ is infinite almost surely. Since $\bigcap\F$ is finite, we therefore have $\A(S_y)\setminus\bigcap\F\ne\varnothing$ almost surely. Because $z_1,z_2,\dots$ enumerate $\X\setminus\bigcap\F$, it follows that, for each fixed $y\in[i]^n$,
    \begin{equation*}
        \lim_{q\to\infty}
        \Pr[\A(S_y)\cap\{z_1,\dots,z_q\}\ne\varnothing]=1.
    \end{equation*}
    Since there are only finitely many $y\in[i]^n$, we can choose $q_i$ such that, for every $y\in[i]^n$,
    \begin{equation*}
        \Pr[\A(S_y)\cap\{z_1,\dots,z_{q_i}\}\ne\varnothing]
        \ge1-\eta.
    \end{equation*}
    For each $q\in[q_i]$, choose $F_{i,q}\in\F$ that excludes $z_q$, and define $B_i=B_{i-1}\cap\bigcap_{q=1}^{q_i}F_{i,q}$. Since every finite subfamily of $\F$ has infinite intersection, $B_i$ is infinite and belongs to $\CC$. Moreover, $B_i$ excludes $z_1,\dots,z_{q_i}$, so for every $y\in[i]^n$,
    \begin{equation*}
        \Pr[\A(S_y)\subseteq B_i]
        \le
        \Pr[\A(S_y)\cap\{z_1,\dots,z_{q_i}\}=\varnothing]
        \le\eta.
    \end{equation*}

    Iterate the above construction until $B_N$ is constructed. Let $\M$ be the mechanism that, on input $y\in[N]^n$, outputs the least $i\in[N]$ such that $\A(S_y)\not\subseteq B_i$, using $N$ if no such $i$ exists. Changing one coordinate of $y$ changes only one entry of $S_y$, and $\M$ is a post-processing of $\A$. Hence $\M$ is $(\varepsilon,\delta)$-differentially private.

    Fix $y\in[N]^n$, and write $\ell=\min_t y_t$ and $r=\max_t y_t$. For each $t\in[n]$, nestedness gives $a_{y_t,t}\in B_{y_t-1}\subseteq B_{\ell-1}$.
    Thus $\widehat S_y\subseteq B_{\ell-1}$, and Lemma~\ref{lem:lower_closure} implies that $\A(S_y)\subseteq B_{\ell-1}$ almost surely. Since $B_{\ell-1}\subseteq B_i$ for $i<\ell$, no index below $\ell$ can be returned. Hence $\M(y)\ge\ell$ almost surely.

    Since $y\in[r]^n$, the construction gives $\Pr[\A(S_y)\subseteq B_r]\le\eta$. Whenever $\A(S_y)\not\subseteq B_r$, the first failed containment occurs at an index at most $r$. Consequently,
    \begin{align*}
        \Pr[\ell\le\M(y)\le r]
        &=\Pr[\M(y)\le r]\\
        &\ge\Pr[\A(S_y)\not\subseteq B_r]\\
        &=1-\Pr[\A(S_y)\subseteq B_r]\\
        &\ge1-\eta.
    \end{align*}
    Taking $N$ and $\eta$ supplied by Lemma~\ref{lem:lower_sharp_interior_point} yields a contradiction.
\end{proof}

\subsection{Binary Branching}
We compare the ranks defined in Section~\ref{sec:dimension}. The proof of Fact~\ref{fact:upper_rank_monotonicity} also shows that $\rho_k(H)\le\rho_k(K)$ whenever $H,K\in\CC$ and $H\subseteq K$.

\begin{lemma}
\label{lem:lower_intersection}
    Let $k\ge2$ and $s\ge1$, and let $H,K\in\CC$ with $H\subseteq K$ and $\rho_k(H)\le s$. Any $k$-ary nested branching tree with root label $K$ and depth at least $s+1$ has a node $v$ with label $S_v$ such that $\lvert S_v\cap H\rvert<\infty$ and the unique path from the root to $v$ contains at most $s$ edges.
\end{lemma}
\begin{proof}
    Suppose the conclusion fails. Truncate the tree to its first $s+1$ levels. For every remaining node $v$, the set $H\cap S_v$ is infinite and therefore belongs to $\CC$. Replace each label $S_v$ by $H\cap S_v$. The root label becomes $H\cap K=H$, and nesting is preserved. Each intersection of child labels remains finite because it is contained in the original intersection. Thus the resulting tree is a $k$-ary nested branching tree with root label $H$ and depth $s+1$, contradicting $\rho_k(H)\le s$.
\end{proof}

\begin{lemma}
\label{lem:lower_arity}
    For every $k\ge2$ and $K\in\CC$ with $\rho_2(K)<\infty$,
    \begin{equation*}
        \rho_k(K)\le 2^{\rho_2(K)}-1.
    \end{equation*}
\end{lemma}
\begin{proof}
    Let $f(q)=2^q-1$. We prove by induction on $q\ge1$ that $\rho_2(K)\le q$ implies $\rho_k(K)\le f(q)$. If $q=1$ and a $k$-ary tree rooted at $K$ has two levels, choose a minimal collection $A_1,\dots,A_r$ of its child labels with finite intersection. Every $A_i$ is infinite, so $r\ge2$ and $B=\bigcap_{i=2}^r A_i$ is an infinite member of $\CC$. The labels $A_1,B\subseteq K$ have finite intersection and give a binary tree with two levels, a contradiction. Therefore, $\rho_k(K)\le 1 = f(1)$.

    Suppose the claim holds at $q-1$, let $s=f(q-1)$, and assume that $\rho_2(K)\le q$ while a $k$-ary tree rooted at $K$ has depth $2s+2=f(q)+1$. Each child roots a tree of depth $2s+1>s$, so the induction hypothesis and rank monotonicity show that its binary rank is exactly $q$. Choose a minimal collection $A_1,\dots,A_r$ of child labels with finite intersection. If $r=2$, attaching two binary trees of depth $q$ below $K$ gives binary depth $q+1$, a contradiction.

    Thus $r\ge3$. For $j=2,\dots,r$, let $Q_j=\bigcap_{i=j}^r A_i$. Minimality makes every $Q_j$ infinite, hence a member of $\CC$. Since $A_1\cap Q_2$ is finite and $\rho_2(A_1)=q$, we have $\rho_2(Q_2)\le q-1$, and therefore $\rho_k(Q_2)\le s$ by the induction hypothesis.

    We prove $\rho_k(Q_j)\le s$ for $j=2,\dots,r$ by induction on $j$. The case $j=2$ was established above. Suppose $\rho_k(Q_j)\le s$ for some $2\le j<r$. Apply Lemma~\ref{lem:lower_intersection} to $Q_j\subseteq A_j$ and the subtree with root label $A_j$. It yields a node at most $s$ edges below this root, labeled $B_j\subseteq A_j$, with $\lvert B_j\cap Q_j\rvert<\infty$. Its subtree has depth at least $2s+1-s=s+1$, so the induction hypothesis at $q-1$ and monotonicity give $\rho_2(B_j)=q$.

    Since $B_j\subseteq A_j$ and $Q_j=A_j\cap Q_{j+1}$, we have $B_j\cap Q_{j+1}=B_j\cap Q_j$, which is finite. If $\rho_2(Q_{j+1})\ge q$, attaching binary trees of depth $q$ with root labels $B_j$ and $Q_{j+1}$ below a new root labeled $K$ gives depth $q+1$, a contradiction. Thus $\rho_2(Q_{j+1})\le q-1$, and the induction hypothesis at $q-1$ gives $\rho_k(Q_{j+1})\le s$.

    Ultimately, we have $\rho_k(Q_r)\le s$. But $Q_r=A_r$, whose subtree has depth $2s+1>s$, a contradiction.
\end{proof}

\begin{corollary}[Rank comparison]
\label{cor:lower_rank_comparison}
    If $\h(\C)<\infty$ and $\nbd_2(\C)<\infty$, then
    \begin{equation*}
        \nbd(\C)\le 2^{\nbd_2(\C)}-1.
    \end{equation*}
\end{corollary}
\begin{proof}
    Let $h=\h(\CC)=\h(\C)$, where the equality follows from Lemma~\ref{lem:Helly_pass}. If $h=1$, any internal node would have a child with a finite label, which is impossible. Thus both $\nbd(\C)$ and $\nbd_2(\C)$ equal $1$, and the bound holds.

    Assume $h\ge2$. At each internal node, the child labels belong to $\CC$ and have finite intersection. We can retain at most $h$ children whose labels still have finite intersection. Repeating this at every retained internal node preserves the root label and depth. Thus, for every $K\in\CC$, Lemma~\ref{lem:lower_arity} gives $\rho(K)=\rho_h(K)\le2^{\rho_2(K)}-1\le2^{\nbd_2(\C)}-1$. Taking the supremum over $K\in\CC$ gives the result.
\end{proof}

We next bound the depth of a binary tree by reducing to private interior point. We first record the quantitative lower bound for this problem. The proof is given in Appendix~\ref{app:interior_point_proofs} for completeness.

\begin{lemma}[Interior point lower bound~\cite{bun2015differentially}]
\label{lem:lower_quantitative_interior_point}
    Let $m\ge1$ and $N\ge2$, and let $0\le\varepsilon\le1$ and $0\le\delta\le1/(12(m+1))$. Suppose $\M:[N]^m\to[N]$ is $(\varepsilon,\delta)$-differentially private and satisfies
    \begin{equation*}
        \Pr\left[\min_{i\in[m]}y_i\le\M(y)\le\max_{i\in[m]}y_i\right]\ge\frac23
    \end{equation*}
    for every $y\in[N]^m$. Then $m=\Omega(\log^*N)$.
\end{lemma}

\begin{lemma}
\label{lem:lower_binary}
    Let $\A$ be an $(\varepsilon,\delta)$-differentially private uniform generation algorithm for $\C$ with sample size $n\ge2$, where $0\le\varepsilon\le1$ and $\delta\le1/(12n)$. If $\C$ has a binary nested branching tree of depth $d$, then $n=\Omega(\log^* d)$.
\end{lemma}
\begin{proof}
    Let $s=\lfloor(d-1)/4\rfloor$. We may assume $s\ge2$, since otherwise $d\le8$ and the bound follows from $n\ge2$. Truncate the binary tree to depth $4s+1$, and label the two outgoing edges at each internal node by $0$ and $1$. Group every four consecutive edges into one edge labeled by a symbol of $B=\{0,1\}^4$. The resulting tree has depth $s+1$. For each word $y\in B^{\le s}$, denote the corresponding node label by $K_y$.

    For $p\in B^{<s}$ and distinct $a,a'\in B$, the paths to $pa$ and $pa'$ first diverge at two children whose labels have finite intersection. Since $K_{pa}$ and $K_{pa'}$ are contained in these respective labels, we have $\lvert K_{pa}\cap K_{pa'}\rvert<\infty$. Since all leaf labels are infinite, choose distinct points $u_{i,y}\in K_y$ for $i\in\{0,\dots,n-1\}$ and $y\in B^s$.

    Given $x=(x_1,\dots,x_{n-1})\in\{0,\dots,s-1\}^{n-1}$, choose a reference word $Y_0\in B^s$ and symbols $U_{i,j}\in B$ for $i\in[n-1]$ and $j\in[s]$, uniformly over all such choices. Let $Y_i$ agree with $Y_0$ on its first $x_i$ symbols, and set $Y_i^{(j)}=U_{i,j}$ for $j>x_i$. Run $\A$ on $S_x=(u_{0,Y_0},u_{1,Y_1},\dots,u_{n-1,Y_{n-1}})$, whose entries are distinct. For $0\le r\le s$, write $Y_0|r=(Y_0^{(1)},\dots,Y_0^{(r)})$ for the prefix of length $r$ of $Y_0$, with $Y_0|0=()$. Define $\M(x)$ as the largest $r\in\{0,\dots,s-1\}$ such that $\A(S_x)\subseteq K_{Y_0|r}$, taking $\M(x)=0$ if no such $r$ exists.

    All input points belong to $K_{Y_0|\min_i x_i}$, because every $Y_i$ shares its first $\min_i x_i$ symbols with $Y_0$. Lemma~\ref{lem:lower_closure} therefore implies that $\A(S_x)$ is almost surely an infinite subset of this label. Consequently, $\M(x)\ge\min_i x_i$ almost surely.

    Fix the encoding seed $\theta=(Y_0,(U_{i,j})_{i,j})$. Changing one coordinate of $x$ changes at most one entry of $S_x$, and the decoder depends only on $Y_0$ and $\A(S_x)$. Thus differential privacy and post-processing give the required guarantee for each fixed seed. Since the seed distribution does not depend on $x$, averaging over it shows that $\M$ is $(\varepsilon,\delta)$-differentially private.

    To bound $\M(x)$ from above, let $w=\max_i x_i$ and condition on all encoding randomness except $V=Y_0^{(w+1)}$, which remains uniform on $B$. Since $x_i\le w$, none of $Y_1,\dots,Y_{n-1}$ depends on $V$. Changing $V$ therefore changes only the first entry of $S_x$. For $v\in B$, let $D_v$ be the dataset obtained by setting $V=v$, and let $E_v$ consist of the infinite subsets of $K_{(Y_0|w)v}$. Thus $\A(D_v)\in E_v$ means that $\A(D_v)$ is infinite and contained in this label. The sets $E_v$ are pairwise disjoint because the corresponding labels have finite intersections.

    Under this conditioning, $\M(x)>w$ requires $\A(S_x)\subseteq K_{(Y_0|w)V}$, so
    \begin{equation*}
        \Pr[\M(x)>w]\le\frac1{16}\sum_{v\in B}\Pr[\A(D_v)\in E_v].
    \end{equation*}
    Applying privacy to $D_v,D_{v'}$ and averaging over both indices gives
    \begin{equation*}
        \Pr[\M(x)>w]\le\frac{e^\varepsilon}{16^2}\sum_{v,v'\in B}\Pr[\A(D_{v'})\in E_v]+\delta.
    \end{equation*}
    For each fixed $v'$, the pairwise disjointness of the $E_v$ gives $\sum_{v\in B}\Pr[\A(D_{v'})\in E_v]\le1$. Hence $\Pr[\M(x)>w]\le e^\varepsilon/16+\delta<1/3$. Averaging over the remaining randomness preserves this bound, so $\Pr[\min_i x_i\le\M(x)\le\max_i x_i]>2/3$.

    Apply Lemma~\ref{lem:lower_quantitative_interior_point} with $m=n-1$ and $N=s$ (after shifting the ordered domain by one) to obtain $n-1=\Omega(\log^*s)$. Since $s=\lfloor(d-1)/4\rfloor$, we have $\log^*d\le\log^*s+O(1)$, and hence $n=\Omega(\log^*d)$.
\end{proof}

\subsection{Wrapping Up}
We prove our main lower bound by combining the results proved in this section.
\begin{proof}[Proof of Theorem~\ref{thm:lower_full}]
    Lemma~\ref{lem:lower_closure_dimension} gives $n\ge\cd(\C)+1$. Since $\delta\le1/(12n)<1/n$, Lemma~\ref{lem:lower_helly} gives $\h(\C)\le n$.

    It remains to prove $n=\Omega(\log^*\nbd(\C))$. If $\h(\C)=1$, then $\nbd(\C)=1$, and this bound is immediate. Assume $\h(\C)\ge2$, so $n\ge2$.

    Lemma~\ref{lem:lower_binary} rules out binary nested branching trees of arbitrarily large depth, so $\nbd_2(\C)<\infty$. Applying the same lemma at depth $\nbd_2(\C)$ gives $\log^*\nbd_2(\C)=O(n)$.

    Corollary~\ref{cor:lower_rank_comparison} now gives $\nbd(\C)\le2^{\nbd_2(\C)}-1$, and hence
    \begin{equation*}
        \log^*\nbd(\C)\le1+\log^*\nbd_2(\C)=O(n),
    \end{equation*}
    which gives the remaining bound.
\end{proof}

\section{Discussion}
\label{sec:discussion}

We characterize uniform language generation under approximate differential privacy by proving that a language class admits a private uniform generator if and only if both its closure dimension and nested branching dimension are finite. Below, we discuss the connection to private learning and the gap between our upper and lower bounds.

\paragraph*{Connection to private learning.}

Uniform generatability and private learnability are incomparable because their characterizing dimensions, closure dimension and Littlestone dimension, are incomparable~\cite{raman2025generation,alon2019private,bun2020equivalence}. Even among uniformly generatable classes, private learnability and private uniform generatability are incomparable. An intuitive explanation is that nested branching trees are constructed using $\CC$, whereas Littlestone dimension is defined using $\C$. Finite $\ld(\CC)$ implies finite $\nbd(\C)$. Consequently, for a uniformly generatable class $\C$, the private learnability of $\CC$ implies the private uniform generatability of $\C$, but the converse fails. Examples and proofs appear in Appendix~\ref{app:private_learning}.

\paragraph*{Remaining gap.}

Our upper and lower bounds leave a substantial gap in sample complexity. The statements in the introduction suppress the dependence on the finite-intersection Helly number $\h(\C)$. The privacy term in the upper bound of Theorem~\ref{thm:upper_private_selection} has $\h(\C)$ as a leading multiplicative factor, while Theorem~\ref{thm:lower_full} gives the lower bound $n\ge\h(\C)$. The examples in Appendix~\ref{app:helly_nbd_examples} illustrate two extreme cases. For classes with Helly number $2$ and growing nested branching dimension, the upper bound has a privacy term smaller by a factor of order $\nbd(\C)$ than the version stated in the introduction, although its dependence on $\nbd(\C)$ remains exponential. When the Helly number grows linearly with the nested branching dimension, the lower bound is already linear in $\nbd(\C)$, strengthening the $\Omega(\log^*\nbd(\C))$ bound stated in the introduction. Determining the optimal joint dependence of sample complexity on the Helly number and nested branching dimension remains open.


\section*{AI Disclosure}
GPT 6 Astra was used to assist with language editing, grammatical corrections, and the construction of the examples in Appendices~\ref{app:closure_nbd_incomparability},~\ref{app:incomparability_original}, and~\ref{app:helly_nbd_examples}. The authors' initial proof of Lemma~\ref{lem:lower_helly} used the axiom of countable choice over \(\mathcal P(\mathbb N)\). GPT 6 Astra subsequently proposed an alternative proof that does not rely on this axiom. The authors reviewed the AI-assisted contributions and take responsibility for the correctness and final presentation of the mathematical results.

\bibliographystyle{alpha}
\bibliography{references}

\newpage
\appendix
\section{Proof of Lemma~\ref{lem:closure_nbd_incomparability}}
\label{app:closure_nbd_incomparability}

The following two lemmas imply Lemma~\ref{lem:closure_nbd_incomparability}.

\begin{lemma}
\label{lem:app_finite_cd_infinite_nbd}
    On $\X_1=\mathbb Q$, let
    \begin{equation*}
        \C_1=\{(a,b)\cap\mathbb Q:a,b\in\mathbb Q,\ a<b\}.
    \end{equation*}
    Then $\C_1$ is countable, $\cd(\C_1)=1$, and $\nbd(\C_1)=\infty$.
\end{lemma}
\begin{proof}
    For any non-empty finite $T\subseteq\mathbb Q$, every language containing $T$ contains $[\min T,\max T]\cap\mathbb Q$, and every rational point outside this interval can be excluded by such a language. Hence $\langle T\rangle_{\C_1}=[\min T,\max T]\cap\mathbb Q$. This closure is finite exactly when $T$ is a singleton, so $\cd(\C_1)=1$.

    Every member of $\C_1$ contains two disjoint members of $\C_1$. Repeating this construction yields binary nested branching trees of arbitrarily large depth, so $\nbd(\C_1)=\infty$.
\end{proof}

\begin{lemma}
\label{lem:app_infinite_cd_finite_nbd}
    On $\X_2=\mathbb N^2$, define
    \begin{equation*}
        L_n=(\{0\}\times[n])\cup(\{n\}\times\mathbb N),
        \qquad
        \C_2=\{L_n:n\ge1\}.
    \end{equation*}
    Then $\C_2$ is countable, $\cd(\C_2)=\infty$, and $\nbd(\C_2)=1$.
\end{lemma}
\begin{proof}
    For distinct $n,m$, we have $L_n\cap L_m=\{0\}\times[\min\{n,m\}]$. For $T_n=\{0\}\times[n]$, both $L_n$ and $L_{n+1}$ contain $T_n$, so $\langle T_n\rangle_{\C_2}\subseteq L_n\cap L_{n+1}=T_n$. Thus $T_n$ has finite closure and size $n$, proving $\cd(\C_2)=\infty$.

    Every subfamily with at least two members has finite intersection, so $(\C_2)^\cap_\infty=\C_2$. Moreover, $(m,1)\in L_m\setminus L_n$ whenever $m\ne n$, so $L_m\not\subseteq L_n$. Consequently, every child of a node labeled $L_n$ would also be labeled $L_n$, making the intersection of its child labels infinite. Therefore, $\nbd(\C_2)=1$.
\end{proof}

\section{Interior Point Lower Bounds}
\label{app:interior_point_proofs}

We prove Lemmas~\ref{lem:lower_sharp_interior_point} and~\ref{lem:lower_quantitative_interior_point} using interior point fingerprinting results from Appendix B of the full version of~\cite{bun2015differentially}. We refer the reader to their work for the notation and terminology of interior point fingerprinting codes.

\begin{proof}[Proof of Lemma~\ref{lem:lower_sharp_interior_point}]
    If $n=1$, take $N=2$ and $0<\eta<(1-\delta)/(1+e^\varepsilon)$. Accuracy on both singleton inputs and privacy for $\{1\}$ would give $1-\eta\le e^\varepsilon\eta+\delta$, a contradiction.

    For $n\ge2$, choose $\eta,\xi>0$ with $e^\varepsilon\xi+\delta<(1-\eta)/n$. By \cite[Lemma B.4]{bun2015differentially}, take an $n$-user interior point fingerprinting code with perfect completeness and soundness at most $\xi$ on a finite domain $[N]$. Since $\xi<1/n$, necessarily $N\ge2$. Let $C=(c_1,\dots,c_n)$ be its codebook and $\mathrm{Trace}$ be its tracer.

    Suppose $\M$ succeeds with probability at least $1-\eta$ on every input. Perfect completeness and averaging over the users give some $i\in[n]$ with $\Pr[\mathrm{Trace}(\M(C))=i]\ge(1-\eta)/n$.

    Replace $c_i$ by $c_j$ for any $j\ne i$ and obtain $C^{(i)}$. This dataset uses only the codewords of the coalition $[n]\setminus\{i\}$, so soundness gives $\Pr[\mathrm{Trace}(\M(C^{(i)}))=i]\le\xi$. Applying privacy conditioned on code generation and tracing randomness, then averaging, yields
    \begin{equation*}
        \frac{1-\eta}{n}\le e^\varepsilon\xi+\delta,
    \end{equation*}
    contradicting the choice of $\eta,\xi$.
\end{proof}

\begin{proof}[Proof of Lemma~\ref{lem:lower_quantitative_interior_point}]
    Let $q=m+1$. By \cite[Lemma B.4]{bun2015differentially}, there is a $q$-user interior point fingerprinting code on $[Q]$ with perfect completeness, soundness at most $1/(33q)$, and $\log^*Q\le q+\log^*(66q^3)=O(q)$. Since $\varepsilon\le1$ and $\delta\le1/(12q)$, Lemma~B.3 of \cite{bun2015differentially} rules out an algorithm with the stated privacy and accuracy guarantees on $[Q]^m$.

    If $N\ge Q$, restrict $\M$ to inputs in $[Q]^m$ and clip its output to $[Q]$. This preserves privacy and every successful output, giving a contradiction. Thus $N<Q$, and $\log^*N=O(q)=O(m)$.
\end{proof}

\section{Connection to the Littlestone Dimension}
\label{app:private_learning}

This appendix establishes the comparisons between Littlestone dimension and nested branching dimension discussed in Section~\ref{sec:discussion}. We view each language as its indicator function. Below we briefly state the definition of Littlestone dimension for language classes. 

Consider a complete binary tree of depth $d$ whose nodes are labeled by points in $\X$. Every $(y_1,\dots,y_d)\in\{0,1\}^d$ determines a unique root-to-leaf path $(v_1,\dots,v_d)$, where $v_1$ is the root and, for each $i\in[d-1]$, $v_{i+1}$ is the left child of $v_i$ if $y_i=0$ and the right child if $y_i=1$. Let $(x_1,\dots,x_d)$ denote the sequence of labels along this path. The tree is said to be shattered by $\C$ if, for every $(y_1,\dots,y_d)\in\{0,1\}^d$ and its corresponding path labels $(x_1,\dots,x_d)$, there exists $L\in\C$ such that $y_i = \mathbb I[x_i\in L]$. The Littlestone dimension of $\C$, denoted by $\ld(\C)$, is the supremum of the depths of complete binary trees shattered by $\C$.

\subsection{Incomparability for the Original Language Class}
\label{app:incomparability_original}

We give two countable language classes, both with closure dimension $0$, witnessing the two directions of incomparability.

\begin{lemma}
\label{lem:app_finite_ld_infinite_nbd}
    On $\X=\mathbb N^2$, let
    \begin{equation*}
        L_i=(\mathbb N\setminus\{i\})\times\mathbb N,
        \qquad \C_1=\{L_i:i\in\mathbb N\}.
    \end{equation*}
    Then $\cd(\C_1)=0$, $\ld(\C_1)=1$, and $\nbd(\C_1)=\infty$.
\end{lemma}
\begin{proof}
    For a non-empty finite $T\subseteq\X$, let $I(T)$ be the set of its first coordinates. The languages containing $T$ are exactly the $L_i$ with $i\notin I(T)$, so $\langle T\rangle_{\C_1}=I(T)\times\mathbb N$, which is infinite. Since $\bigcap\C_1=\varnothing$, we obtain $\cd(\C_1)=0$.

    Since $(1,1)\notin L_1$ and $(1,1)\in L_2$, we have $\ld(\C_1)\ge1$. The $0$-branch at a root labeled $(i,j)$ leaves only $L_i$, which cannot realize a further split. Hence $\ld(\C_1)=1$.

    For every non-empty proper subset $A\subsetneq\mathbb N$, we have
    \begin{equation*}
        A\times\mathbb N
        =\bigcap_{i\in\mathbb N\setminus A}L_i
        \in(\C_1)^\cap_\infty.
    \end{equation*}
    Starting from an infinite proper $A$, recursively split each subset into two disjoint infinite subsets and use their products with $\mathbb N$ as node labels. This gives binary nested branching trees of arbitrary depth, so $\nbd(\C_1)=\infty$.
\end{proof}

\begin{lemma}
\label{lem:app_finite_nbd_infinite_ld}
    Let $A,B,D$ be disjoint countably infinite sets, let $\X=A\cup B\cup D$, and define
    \begin{equation*}
        \C_2=\{B\cup F:F\subseteq A,\ \lvert F\rvert<\infty\}\cup\{D\}.
    \end{equation*}
    Then $\cd(\C_2)=0$, $\nbd(\C_2)=1$, and $\ld(\C_2)=\infty$. Moreover, $(\C_2)^\cap_\infty=\C_2$.
\end{lemma}
\begin{proof}
    Let $T$ be a non-empty finite set contained in some language from $\C_2$. The least language containing $T$ is $D$ if $T\subseteq D$, and $B\cup(T\cap A)$ otherwise. Thus every such closure is infinite. Since $B,D\in\C_2$ and $B\cap D=\varnothing$, we obtain $\cd(\C_2)=0$.

    Intersecting a non-empty family of languages $B\cup F$ gives $B\cup G$ for some finite $G\subseteq A$, while $D$ is disjoint from every other language. Thus $(\C_2)^\cap_\infty=\C_2$. Every child label below $B\cup F$ contains $B$, and every child label below $D$ equals $D$. In either case, the intersection of the child labels is infinite, so $\nbd(\C_2)=1$.

    For any $m\ge1$, choose distinct $a_1,\ldots,a_m\in A$. The languages $B\cup F$, with $F\subseteq\{a_1,\ldots,a_m\}$, realize all binary labelings of these points. Thus the complete binary tree querying these points in order is shattered. Since $m$ is arbitrary, $\ld(\C_2)=\infty$.
\end{proof}

\subsection{Passing to the Infinite-Intersection Closure}

\begin{lemma}
\label{lem:app_ld_controls_nbd}
    If $\ld(\CC)=\ell<\infty$, then
    \begin{equation*}
        \nbd(\C)\le 2^{\ell+1}-1.
    \end{equation*}
    In particular, finite Littlestone dimension of $\CC$ implies finite nested branching dimension of $\C$.
\end{lemma}
\begin{proof}
    Consider a nested branching tree for $\C$ of depth at least $2^{\ell+1}$. Every internal node has a child whose label is a proper subset of its parent label. Otherwise, all child labels would equal the infinite parent label, violating the branching condition. Following such children gives a strict chain of $2^{\ell+1}$ members of $\CC$.

    Query a point in the difference between the two middle sets of this chain. The two labels split the chain into equal halves. Repeating this construction within each half gives a complete binary tree of depth $\ell+1$, with every root-to-leaf path realized by a member of the chain. This contradicts $\ld(\CC)=\ell$. Therefore, $\nbd(\C)\le2^{\ell+1}-1$.
\end{proof}

\section{Helly Number and Nested Branching Dimension}
\label{app:helly_nbd_examples}

The following two families of finite language classes realize the regimes discussed in Section~\ref{sec:discussion}. Both have closure dimension $0$. In the first family, the Helly number stays equal to $2$ while the nested branching dimension grows without bound. In the second family, the Helly number equals the nested branching dimension plus one.

\subsection{Constant Helly Number}

\begin{lemma}
\label{lem:app_constant_helly_growing_nbd}
    Fix an integer $q\ge1$ and let $\X_q=\{0,1\}^q\times\mathbb N$. For each binary word $s$ of length at most $q$, including the empty word, define
    \begin{equation*}
        L_s=\{(w,j)\in\X_q:s\text{ is a prefix of }w\},
        \qquad
        \C_{\mathrm{bin},q}=\{L_s:s\in\{0,1\}^{\le q}\}.
    \end{equation*}
    Then $\cd(\C_{\mathrm{bin},q})=0$, $\h(\C_{\mathrm{bin},q})=2$, and $\nbd(\C_{\mathrm{bin},q})=q+1$.
\end{lemma}
\begin{proof}
    Every $L_s$ is infinite, and any two are nested or disjoint. A non-empty subfamily with no disjoint pair is a finite chain, so its intersection is its smallest member. Thus $(\C_{\mathrm{bin},q})^\cap_\infty=\C_{\mathrm{bin},q}$, and every subfamily with finite intersection contains a disjoint pair. Since $L_0\cap L_1=\varnothing$, we obtain $\h(\C_{\mathrm{bin},q})=2$.

    For a non-empty finite $T\subseteq\X_q$, let $p$ be the longest common prefix of its first coordinates. Then $L_p$ is the least language containing $T$, so $\langle T\rangle_{\C_{\mathrm{bin},q}}=L_p$ is infinite. Since the whole class has empty intersection, $\cd(\C_{\mathrm{bin},q})=0$.

    The complete binary tree with node labels $L_s$ and child labels $L_{s0},L_{s1}$ is a nested branching tree of depth $q+1$, giving $\nbd(\C_{\mathrm{bin},q})\ge q+1$. Conversely, the branching condition forces every internal node of any nested branching tree for the class to have a child with a strictly smaller label. Following such children gives a root-to-leaf chain in which the lengths of the indexing words strictly increase. These lengths lie between $0$ and $q$, so the tree has depth at most $q+1$. Therefore, $\nbd(\C_{\mathrm{bin},q})=q+1$.
\end{proof}

\subsection{Linear Helly Number}

\begin{lemma}
\label{lem:app_linear_helly_nbd}
    Fix an integer $k\ge2$ and let $\X_k=[k]\times\mathbb N$. Define
    \begin{equation*}
        L_i=([k]\setminus\{i\})\times\mathbb N,
        \qquad
        \C_{\mathrm{omit},k}=\{L_i:i\in[k]\}.
    \end{equation*}
    Then $\cd(\C_{\mathrm{omit},k})=0$, $\h(\C_{\mathrm{omit},k})=k$, and $\nbd(\C_{\mathrm{omit},k})=k-1$.
\end{lemma}
\begin{proof}
    For every non-empty $I\subseteq[k]$, we have $\bigcap_{i\in I}L_i=([k]\setminus I)\times\mathbb N$. This intersection is finite exactly when $I=[k]$. Hence $\h(\C_{\mathrm{omit},k})=k$, and
    \begin{equation*}
        (\C_{\mathrm{omit},k})^\cap_\infty
        =\{A\times\mathbb N:\varnothing\subsetneq A\subsetneq[k]\}.
    \end{equation*}

    Let $T$ be a non-empty finite set contained in some language from the class, and let $J\subsetneq[k]$ be its set of first coordinates. The languages containing $T$ are exactly those with indices outside $J$, so $\langle T\rangle_{\C_{\mathrm{omit},k}}=J\times\mathbb N$ is infinite. Since the whole class has empty intersection, $\cd(\C_{\mathrm{omit},k})=0$.

    Construct a tree with root label $[k-1]\times\mathbb N$. At each node labeled $A\times\mathbb N$ with $\lvert A\rvert\ge2$, use the child labels $(A\setminus\{i\})\times\mathbb N$ for $i\in A$, and stop at singleton index sets. The child labels have empty intersection, and every root-to-leaf path contains $k-1$ nodes. Hence $\nbd(\C_{\mathrm{omit},k})\ge k-1$.

    Conversely, in any nested branching tree for the class, every internal node has a child with a strictly smaller label. Following such children gives a strict chain of non-empty proper subsets of $[k]$, containing at most $k-1$ members. Thus $\nbd(\C_{\mathrm{omit},k})\le k-1$, proving equality.
\end{proof}

\end{document}